\documentclass{article}

\usepackage[T1]{fontenc}

\usepackage[margin=1in]{geometry}

\usepackage{tikz}
\usepackage{pgfplots}
\usetikzlibrary{calc, arrows.meta, decorations.pathreplacing}

\usepackage{amsmath,amssymb,amsthm}
\usepackage{graphicx}
\usepackage{booktabs}
\usepackage{xcolor}
\usepackage[numbers,sort&compress]{natbib}
\usepackage[most]{tcolorbox}
\usepackage{mathtools}
\usepackage{multirow}
\usepackage{wrapfig}   %

\graphicspath{{figs/}}

\usepackage[colorlinks=true,linkcolor=blue,citecolor=blue,urlcolor=blue]{hyperref}

\usepackage[capitalize]{cleveref}
\crefname{section}{Sec.}{Secs.}
\Crefname{section}{Section}{Sections}
\Crefname{table}{Table}{Tables}
\crefname{table}{Tab.}{Tabs.}

\colorlet{cvprblue}{blue}

\renewcommand{\paragraph}[1]{\vspace{.5em}\noindent\textbf{#1}}

\newtcolorbox[auto counter]{findingbox}[1][]{ enhanced, breakable, colback=gray!5, colframe=black!70, boxrule=0.6pt, arc=1pt, left=6pt, right=6pt, top=5pt, bottom=5pt, fonttitle=\bfseries, title={Finding~\thetcbcounter}, #1 }

\definecolor{txtblue}{RGB}{52, 120, 246}
\definecolor{visorange}{RGB}{240, 128, 40}
\definecolor{darkslate}{RGB}{38, 50, 56}
\definecolor{goodgreen}{RGB}{0, 150, 120}
\definecolor{badred}{RGB}{215, 60, 60}

\newcommand{\tilew}{0.22}  %
\newcommand{\tileh}{0.46}  %
\newcommand{\txttile}[2]{%
  \begin{scope}[shift={(1.9+0.25*#1, #2)}]
    \fill[txtblue!85, rounded corners=0.6pt] (0,0) rectangle (\tilew,\tileh);
    \draw[white, line width=0.5pt] (0.05,0.31) -- (0.17,0.31)
                                   (0.05,0.23) -- (0.17,0.23)
                                   (0.05,0.15) -- (0.13,0.15);
  \end{scope}}
\newcommand{\captile}[2]{%
  \begin{scope}[shift={(1.9+0.25*#1, #2)}]
    \fill[visorange!90, rounded corners=0.6pt] (0,0) rectangle (\tilew,\tileh);
    \fill[white] (0.03,0.09) -- (0.09,0.27) -- (0.13,0.18) -- (0.16,0.23) -- (0.19,0.09) -- cycle;
    \fill[white] (0.155,0.35) circle (0.025);
  \end{scope}}

\newcommand{\Dpt}{D_{\mathrm{pt}}}                 %
\newcommand{\Spt}{S_{\mathrm{pt}}}                 %
\newcommand{\Dcpt}{D_{\mathrm{cpt}}}               %
\newcommand{\Scpt}{S_{\mathrm{cpt}}}               %

\DeclareMathOperator*{\argmin}{arg\,min}

\title{Scaling Laws for Vision-Language Fusion}

\author{Seng Pei Liew \and Teppei Suzuki \and Seiya Tokui}
\date{Noetra Corp.}

\begin{document}
\maketitle
\begin{abstract}
Vision-language models (VLMs) have emerged as central foundation models combining multimodal perception with text comprehension.
While VLMs are traditionally built via late fusion by continued pre-training an off-the-shelf language model on multimodal (image-text) data, mixing modalities from the start of pre-training (early fusion) has recently gained traction.
However, the fundamental trade-offs between these paradigms remain poorly understood, resulting in expensive and heuristic-driven training recipes.
In this paper, we systematically analyze the learning dynamics of these multimodal pre-training processes, showing that validation losses during vision-language fusion can be modeled by separating the base learning trajectory, where different data types contribute at distinct rates, from a distribution shift driven by the competition between modalities.
We derive a scaling law that unifiedly describes these dynamics, enabling accurate loss predictions across mixture ratios, training steps, and model sizes.
Crucially, our law characterizes the trade-off between early and late fusion: under a fixed data budget, early fusion preserves text performance while late fusion favors vision performance, yielding a trade-off that increasingly tilts toward early fusion as model size scales.
Extensive experiments confirm that our scaling law reliably captures multimodal learning dynamics and governs fusion decisions.
\end{abstract}

\section{Introduction}
\label{sec:intro}

Vision-language models (VLMs), combining text comprehension with visual capabilities, have seen rapid growth and adoption in recent years \citep{alayrac2022flamingo, li2023blip, liu2023llava,team2023gemini}.
Traditionally, VLMs are constructed by augmenting a pre-trained LLM with a pre-trained visual encoder (e.g., CLIP \citep{radford2021learning}), and further pre-training it on multimodal data, specifically image-text data (\textit{late fusion}) \citep{koh2023grounding,lin2024vila,mckinzie2024methods}.
More recently, the community has begun to explore training VLMs from scratch or in the early stages, by mixing all modalities in the beginning (\textit{early fusion} or \textit{native multimodal pre-training}) \cite{team2026kimi,wu2026scaling,mistralai2026mistrallarge4}. 
These two paradigms represent contrasting approaches to \textit{fusion timing}, differing in when multimodal data are injected during pre-training.

Given the prohibitive computational cost of large-scale multimodal pre-training, accurately forecasting final performance and resource efficiency prior to execution has become essential.
To this end, scaling laws \citep{6796972,hestness2017deep,hestness2019beyond,rosenfeld2019constructive}, originally studied at large scales in language modeling \citep{kaplan2020scaling,hoffmann2022chinchilla}, have recently been extended to multimodal domains, where a growing body of work has sought to characterize scaling behaviors under this paradigm, providing valuable guidelines for allocating compute, balancing model parameters, and tuning pre-training mixtures \citep{aghajanyan2023scaling,wu2026scaling,khandelwal2026scaling}.

However, a unified scaling law capable of characterizing both early and late fusion within a single formulation remains absent, leaving practitioners unable to systematically determine which paradigm to adopt under a given compute or data budget.
Existing formulations treat early and late fusion in isolation and fail to capture how cross-modal interactions and distribution shifts evolve as fusion timing varies.
Consequently, deciding whether to invest expensive pre-training compute into native multimodal or continued late-fusion pre-training remains largely heuristic.

\begin{figure*}[th]
\centering
\resizebox{\textwidth}{!}{%
\begin{tikzpicture}[
    font=\sffamily, >=Stealth,
    brace/.style={decorate, decoration={brace, amplitude=4pt, raise=1pt}, line width=0.7pt, darkslate},
    rowlbl/.style={anchor=east, align=right, font=\sffamily\bfseries\small, text=darkslate},
    outc/.style={anchor=west, font=\sffamily\scriptsize, text=darkslate}
]

\node[anchor=west, font=\sffamily\bfseries, text=darkslate] at (0, 3.55)
    {(a) Fusion timing under fixed budget};

\begin{scope}[shift={(-0.9, 3.32)}]
  \txttile{25.0}{0}
  \node[anchor=west, font=\sffamily\tiny, text=darkslate] at (8.4, 0.23) {text};
  \captile{28.4}{0}
  \node[anchor=west, font=\sffamily\tiny, text=darkslate] at (9.25, 0.23) {image--caption};
\end{scope}

\node[rowlbl] at (1.75, 2.45) {Late\\[-2pt]fusion};
\foreach \i in {0,...,7} {\txttile{\i}{2.25}}
\foreach \i in {10,13,16,18,21,23,26,28,30,31} {\txttile{\i}{2.25}}
\foreach \i in {8,9,11,12,14,15,17,19,20,22,24,25,27,29} {\captile{\i}{2.25}}

\draw[brace] (1.9, 2.75) -- (3.87, 2.75)
    node[midway, above=3pt, font=\sffamily\scriptsize\bfseries, text=txtblue] 
    {Text-only PT ($D_{\rm pt}$)};
\draw[brace] (3.9, 2.75) -- (9.87, 2.75)
    node[midway, above=3pt, font=\sffamily\scriptsize\bfseries, text=visorange]
    {Multimodal CPT \textcolor{darkslate!80}{(dense mix, high $R$)}};

\draw[darkslate, line width=1.1pt] (3.885, 2.11) -- (3.885, 2.77);
\draw[<->, darkslate!70, line width=0.6pt] (3.1, 2.05) -- (4.67, 2.05);
\node[anchor=north, font=\sffamily\tiny, text=darkslate] at (3.885, 2.01) {fusion timing};

\node[outc] at (5.0, 1.95) {%
  \textcolor{goodgreen}{\bfseries$\blacktriangledown$ lower vision loss}\quad
  \textcolor{badred}{\bfseries$\blacktriangle$ higher text loss}};

\node[rowlbl] at (1.75, 0.90) {Early\\[-2pt]fusion};
\foreach \i in {0,2,4,5,7,9,11,12,14,16,17,19,21,23,24,26,28,30} {\txttile{\i}{0.70}}
\foreach \i in {1,3,6,8,10,13,15,18,20,22,25,27,29,31} {\captile{\i}{0.70}}

\draw[brace] (1.9, 1.18) -- (9.87, 1.18)
    node[midway, above=3pt, font=\sffamily\scriptsize\bfseries, text=darkslate]
    {Train mixing text \& vision from step 0 \ \textcolor{darkslate!80}{($D_{\rm{pt}}=0$, sparse mix, lowest $R$)}};

\node[outc] at (1.9, 0.36) {%
  \textcolor{goodgreen}{\bfseries$\checkmark$ lowest text loss}\quad
  \textcolor{darkslate}{\bfseries$\approx$ vision deficit that shrinks with $N$}};

\draw[->, darkslate!70, line width=0.7pt] (1.9, 0.08) -- (10.0, 0.08);
\node[anchor=north, font=\sffamily\tiny, text=darkslate!80] at (5.9, 0.04)
    {training tokens \ (fixed total budget $T_{\rm txt}+T_{\rm cap}$, identical in both rows)};

\draw[gray!25, line width=0.8pt] (10.45, -0.2) -- (10.45, 3.65);

\node[anchor=west, font=\sffamily\bfseries, text=darkslate] at (10.65, 3.55)
    {(b) Late fusion's vision edge fades with scale};

\pgfmathsetmacro{\txtGap}{2.9}
\pgfmathsetmacro{\capC}{2889.33}

\begin{axis}[
    at={(11.85cm, 0.45cm)}, anchor=south west,
    width=4.75cm, height=2.55cm, scale only axis=true,
    xmode=log, ymode=log,
    xmin=1.7e8, xmax=5.5e9, ymin=0.02, ymax=10,
    xtick={2.722e8,1.1184e9,3.5239e9}, xticklabels={300M,1B,3.5B},
    ytick=\empty,
    tick label style={font=\sffamily\tiny},
    xlabel={model size $N$ (log)}, ylabel={$|L_{\mathrm{late}} - L_{\mathrm{early}}|$ (log)},
    label style={font=\sffamily\scriptsize},
    ylabel style={yshift=-6pt}, xlabel style={yshift=3pt},
    axis line style={darkslate}, tick style={darkslate},
    xmajorgrids, grid style={gray!15},
    clip=false,
]
\addplot[txtblue, line width=1.3pt, mark=*, mark size=2pt]
    coordinates {(2.722e8, \txtGap) (1.1184e9, \txtGap) (3.5239e9, \txtGap)};
\node[anchor=south, font=\sffamily\tiny\bfseries, text=txtblue]
    at (axis cs:1.0e9, 3.3) {text penalty of late fusion ($\gamma_{txt}\!\approx\!0$)};

\addplot[visorange, line width=1.3pt, mark=*, mark size=2pt, domain=2.722e8:3.5239e9, samples at={2.722e8,1.1184e9,3.5239e9}]
    {\capC*x^(-0.5)};
\node[anchor=north, font=\sffamily\tiny\bfseries, text=visorange]
    at (axis cs:7.5e8, 0.05) {visual gain of late fusion $\propto N^{-0.5}$};

\foreach \n in {2.722e8,1.1184e9,3.5239e9} {
  \edef\tmp{\noexpand\draw[<->, darkslate!55, line width=0.5pt]
    (axis cs:\n, {\capC*\n^(-0.5)*1.2}) -- (axis cs:\n, {\txtGap*0.83});}
  \tmp
}

\node[fill=goodgreen!10, draw=goodgreen!70, rounded corners=2pt, inner sep=1.5pt,
      font=\sffamily\tiny\bfseries, text=goodgreen!70!black, align=center]
      at (axis cs:1.98e9, 0.6) {gap widens\\[-1pt]$\Rightarrow$ early\\[-1pt]fusion favored};
\end{axis}

\end{tikzpicture}}
\vspace{-2mm}
\caption{\textbf{Schematic of early versus late fusion in our fusion timing experiment and their scale-dependent performance trade-off.} 
\textbf{(a)} Late fusion begins with text-only pre-training before switching to dense multimodal continued pre-training (high mixture ratio $R$), inducing a large distribution shift that favors vision loss at the cost of text degradation. 
In contrast, early fusion distributes multimodal data across all steps (lowest $R$), avoiding distribution shift to protect text capability at the expense of visual performance.
\textbf{(b)} Across model size, the text-loss penalty paid by late fusion is essentially independent of model size ($\gamma_{\mathrm{txt}}\approx0$), while its vision-loss advantage shrinks: $\propto N^{-0.5}$, so the trade-off tilts increasingly toward early fusion as models scale (see also \cref{tab:downstream_merged}). Combined with the observation that vision loss reduction does not translate to vision benchmark gains (see \cref{sec:fusion}), early fusion becomes the preferred recipe at scale.}
\label{fig:teaser}
\vspace{-3mm}
\end{figure*}
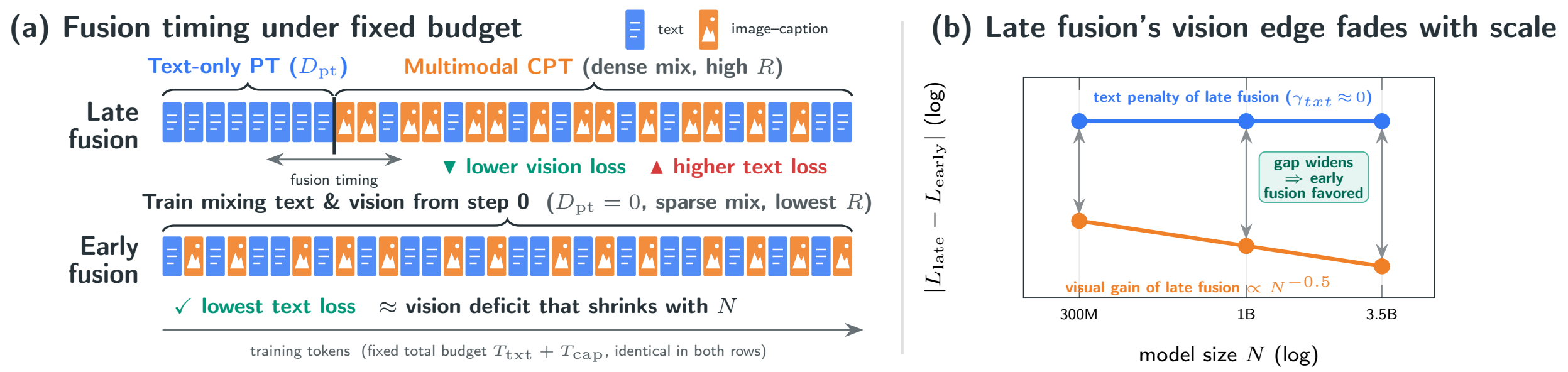

\paragraph{Contributions.}
In this paper, we bridge this gap by establishing a unified scaling law for multimodal pre-training that directly models the trade-offs across different fusion timings/strategies.
At the core of our scaling law is the decomposition of validation loss into two terms: a base learning trajectory where different data types advance progress at distinct exchange rates, and an additive distribution shift governed by competition between modalities.
This formulation spans a continuous spectrum of distribution shifts: in the small-shift regime (\textit{image-text interleaved documents}), cross-modal dynamics resemble standard LLM continued pre-training, whereas in the large-shift regime (\textit{image-caption data}), cross-modal transfer diminishes sharply (\cref{sec:interleaved}).

Critically, our framework reveals that studying fusion timings under a fixed data budget reduces to modeling distribution shift governed by the post-fusion mixture ratio.
At any given model size, this reveals an inherent trade-off: early fusion preserves text performance, whereas late fusion favors vision performance.
Importantly, however, our law reveals that the amplitude of this distribution shift decays with model size (\cref{sec:size}), meaning that the vision advantage of late fusion diminishes at scale, making early fusion increasingly attractive for larger models (\cref{sec:fusion}).
Figure \ref{fig:teaser} illustrates this setup and outcome.

 Furthermore, we find that current vision benchmarks mostly target instruction-following models, counfounding pre-training studies like ours with factors like instruction-following capabilities.
We resolve this by providing new benchmarks to evaluate pre-trained models without requiring instruction-following capabilities and complication due to supervised fine-tuning, enabling a clean evaluation of the scaling law implications beyond validation losses (\cref{sec:downstream_eval}).

\section{Related Work}

\paragraph{Fusion and Scaling Laws.}
Prior fusion timing comparison studies focus on attention structures \citep{han2026physicsmultimodalpretrainingknowledge} and high-level empirical performance \citep{team2026kimi}; they do not provide a quantitative study of the trade-offs between early and late fusion, which is the focus of our work.
Multimodal scaling laws that compare different training paradigms similar to ours are \cite{shukor2025scaling,chen2026from}, but they mainly compare pre-trained visual encoder with encoder-free architecture mixing data modalities altogether from the start.
 We instead use a pre-trained encoder, an approach that has seen adoption by popular open-weight models \cite{team2026qwen3,mistralai2026mistrallarge4,inkling2,team2026kimi}, to study fusion timings.

 \paragraph{Continued Pre-training.}
Late fusion of VLMs may be seen as a form of transfer learning or more generally two-stage learning, of which the scaling behavior in the context of LLMs' final performance has been relatively well-studied \citep{hernandez2021scaling,zhang2024scaling,liew2025scaling,liew2025reusing,akimoto2024m}.
Particularly, continued pre-training, of which late fusion of VLMs is most similar to, has become standard to enhance downstream performance in math, coding, and other domains \cite{sun2020understanding,jang2022continual,gong2022continualpretraining,zan2022cert,yadav2023continualcode}. 
Our scaling-law approach bears similarity to this approach, but with a particular focus on multimodal data,  a different regime of distribution shift than the typical domain transfer setting, and the fusion timing effects.

\section{Preliminaries}

\paragraph{Basic Notation.} 
Throughout this work, we refer to the text-only LLM training as pre-training (PT) and multimodal continued pre-training as continued pre-training (CPT, or late fusion).
Early fusion (or native multimodal pre-training) corresponds to the special case where multimodal training begins from the first step.

\paragraph{Control Variables and Observables.}
We denote the PT and CPT tokens by $\Dpt$ and $\Dcpt$, and their corresponding step counts by $\Spt$ and $\Scpt$.
$R$ represents the multimodal token fraction in the CPT data mixed with text-only data.
The total model size is denoted by $N$.
These are the control variables of our study.

The main observable is the cross-entropy validation loss $L_m$ across modalities $m \in \{{\rm txt, int, cap}\}$, representing the text, image-text interleaved data (interleaved in short), and image-caption data (caption in short) respectively.
We sometimes omit the modality subscript $m$ for notational convenience but restore it (e.g. $L_m$) when comparing or jointly analyzing modalities explicitly.

\paragraph{Fitted Parameters.}
When formulating our scaling laws, parameters are determined via empirical fitting unless stated otherwise.
 By convention, Greek letters ($\alpha, \beta,\gamma,\eta$) denote scaling exponents, while Latin letters ($L_0, A, B$) denote baseline offsets and scaling coefficients.
 All parameters are fitted independently per evaluation modality.

\paragraph{Experimental Setup.}
We train a Gemma-3-style LLM \citep{team2025gemma} (189M to 3.5B) as the backbone of the VLM, and augment it with a SigLIP2 encoder \citep{tschannen2025siglip} to process images.
We use mainly a 300M VLM for the non-model-size scaling studies.
FineWeb-Edu \citep{penedo2023refinedweb} is the text-only data source for PT; OBELICS and OmniCorpus are used as the interleaved source \cite{laurencon2024obelics,li2024omnicorpus}; the caption data in use are from the ReLAION dataset \citep{schuhmann2022laion5b}.
Full hyperparameters and architectural details are in Appendix \ref{app:setup}.

\section{Scaling Vision-Language Fusion}
\label{sec:interleaved}

\begin{figure*}[th]
\centering
\includegraphics[width=\textwidth]{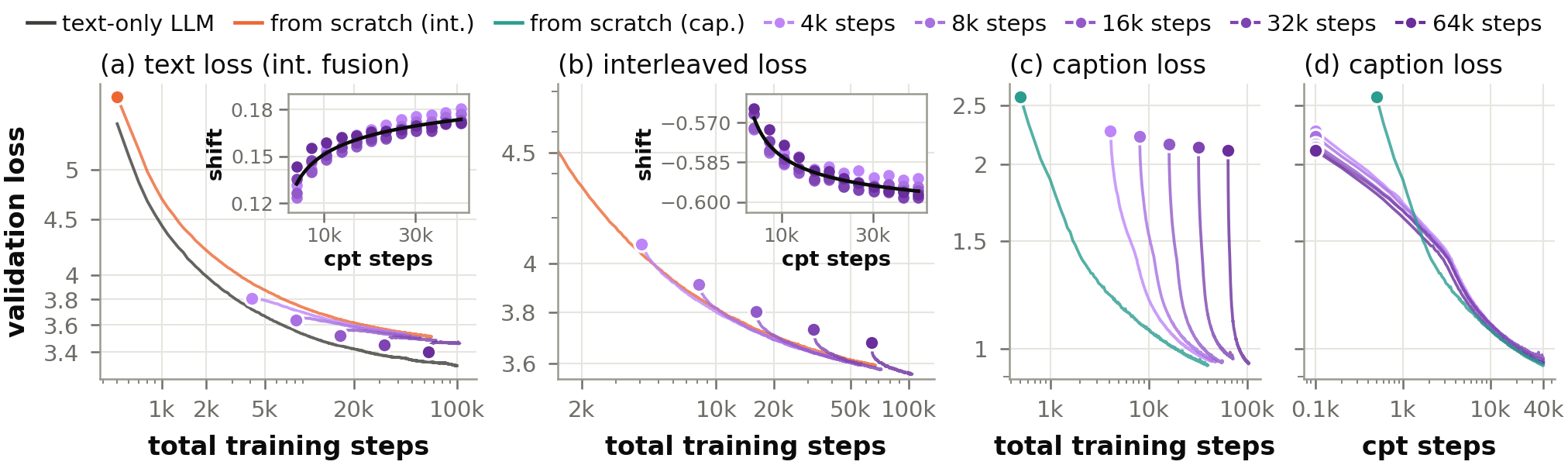}
\caption{ \textbf{Base decomposition and cross-modal exchange rates.}
\textbf{(a,b)} Text and interleaved validation loss curves for interleaved fusion, respectively, across various fusion points $\Dpt$, showing the transition towards the from-scratch trajectory.
 The insets illustrate the shift term $\Delta L$ growing and saturating over CPT steps independent of $\Dpt$. 
 \textbf{(c)} Caption validation loss curves with respect to total training steps across various fusion points $\Dpt$.
 \textbf{(d)} When plotted against CPT steps, caption loss curves collapse onto each other, indicating that text tokens play dimunitive role in caption loss.
 }
\label{fig:curves}
\end{figure*}

In this Section, we study the learning dynamics of multimodal pre-training, considering the fusion with interleaved and caption data \textit{separately}, which represent different regimes of distribution shift, and make comparisons between them.

Interleaved data closely resemble the text-only web corpus used in pre-training, leading to a milder distribution shift.
In practice, however, caption data are the primary source for infusing visual capabilities into VLMs \citep{lin2024vila,mckinzie2024methods,farina2026datacompvlm}.
As captions consist of short alt-texts rather than curated documents, they introduce a pronounced distribution shift during CPT.

\subsection{Base Decomposition and Exchange Rates}
\label{sec:curves}
To establish basic understanding of how validation loss behaves before attempting to formulate the scaling laws, we perform PT on the LLM for various $\Dpt$, and perform CPT using interleaved or caption data ($R=1$) on these checkpoints.
To gain understanding of the behavior, we also pre-train the VLM with interleaved/caption data from scratch and plot its curve, as well as the text-only LLM loss curve.
These serve as control curves for comparison with the true CPT curves.

\paragraph{Observation.}
When multimodal data is introduced during CPT, validation loss deviates from the text pre-training trajectory before settling into a power-law descent. 
On interleaved data, CPT curves transition smoothly from the pre-training baseline toward the multimodal descent (Figure~\ref{fig:curves}\textcolor{cvprblue}{a}).
 The \textit{distribution shift} $\Delta L$ (loss curve minus the base power law) saturates over CPT steps with negligible dependence on $\Dpt$ (Figure~\ref{fig:curves}\textcolor{cvprblue}{a}, inset), motivating the additive decomposition into a base learning trajectory and a transient shift term.
\begin{equation}
L = L_0 + A\, \mathcal{D}_{\text{base}}^{-\alpha} + \Delta L (\Dcpt)
\end{equation}
with $\mathcal{D}_{\text{base}}=\Dpt+\Dcpt$ the \textit{base term}, $L_0$ the irreducible loss, $A$ the scaling coefficient, and $\alpha$ the scaling exponent, all treated as empirical fitting parameters.
These loss behaviors are consistent with those described in the LLM CPT literature (e.g., \cite{ibrahim2024simple}).

 On the other hand, caption data show different behavior.
 As shown in Figure~\ref{fig:curves}\textcolor{cvprblue}{b}, increasing text pre-training $\Dpt$ provides negligible improvement for caption loss. 
 When plotted against CPT steps $\Scpt$ (Figure~\ref{fig:curves}\textcolor{cvprblue}{c}), runs across different $\Dpt$ collapse onto the from-scratch trajectory, indicating that text tokens do not buy much caption loss improvement.

 \paragraph{Formulation.}
 To capture this, we modify $\mathcal{D}_{\text{base},m}$ via a \textit{cross-modal exchange rate} $w_m \in [0, \infty)$, where learning progresses at full strength on native tokens but is modified by $w_m$ on out-of-domain tokens.
 Furthermore, as in practice, multimodal CPT blends multimodal data with text at a multimodal token fraction $R \in [0, 1]$, a total CPT budget of $\Dcpt$ tokens is partitioned into $R \Dcpt$ multimodal tokens and $(1 - R) \Dcpt$ text tokens.
 Summing these contributions leads us to propose the following base term for caption/interleaved and text losses:
\begin{equation}
\label{eq:dbase}
\begin{aligned}
\mathcal{D}_{\text{base},\text{cap/int}} &= w_{\text{cap/int}} \Dpt + \big(R + w_{\text{cap/int}}(1 - R)\big) \Dcpt, \\
\mathcal{D}_{\text{base},\text{txt}} &= \Dpt + \big(w_{\text{txt}} R + (1 - R)\big) \Dcpt.
\end{aligned}
\end{equation}

\begin{wraptable}{l}{0.5\textwidth}
\centering
  \footnotesize
\setlength{\tabcolsep}{4pt}
\begin{tabular}{lcccc}
\toprule
 & \multicolumn{2}{c}{Interleaved fusion} & \multicolumn{2}{c}{Caption fusion} \\
\cmidrule(lr){2-3}\cmidrule(lr){4-5}
 & $L_{\rm int}$ & $L_{\rm txt}$ & $L_{\rm cap}$ & $L_{\rm txt}$ \\
\midrule
$L_0$          & 3.818 & 3.244 & 1.392 & 3.243 \\
$A$            & 90.8  & 89.6  & 694   & 84.5  \\
$\alpha$       & 0.627 & 0.630 & 0.894 & 0.622 \\
$w$            & 1.34  & 0.765 & 0.013 & 0.438 \\
$B_0$          & 0.416 & 0.182 & 0.501 & 0.741 \\
$k_m$          & 36    & 0.185 & 1.95  & 0.105 \\
$\tau$ (steps) & 339   & 7.4   & 1 & 32.8 \\
$\beta$        & 0.170 & 0.299 & 0.353 & 0.463 \\
\midrule
fit RMSE       & 0.0022 & 0.0018 & 0.0099 & 0.0017 \\
held-out RMSE  & 0.0031 & 0.0015 & 0.0084 & 0.0036 \\
\bottomrule
\end{tabular}
\caption{Parameter fits of Equation~\eqref{eq:unified}. $\Dpt$, $\Dcpt$ and $\tau$ are in steps of $2^{20}$ tokens.}
\label{tab:params}
\end{wraptable}

As shown later in the full fit (Table~\ref{tab:params}), $w_{\text{int}} \approx 1.34$, confirming that text and interleaved tokens are near-substitutable in the low-shift regime.
 In contrast, $w_{\text{cap}} \approx 0.013$, explaining why caption learning progresses mainly on caption tokens ($R\Dcpt$) rather than total steps.
\subsection{Mixture Shift and Fractional Competition}
\label{sec:shift_and_competition}
\begin{figure*}[th]
\centering
\includegraphics[width=\textwidth]{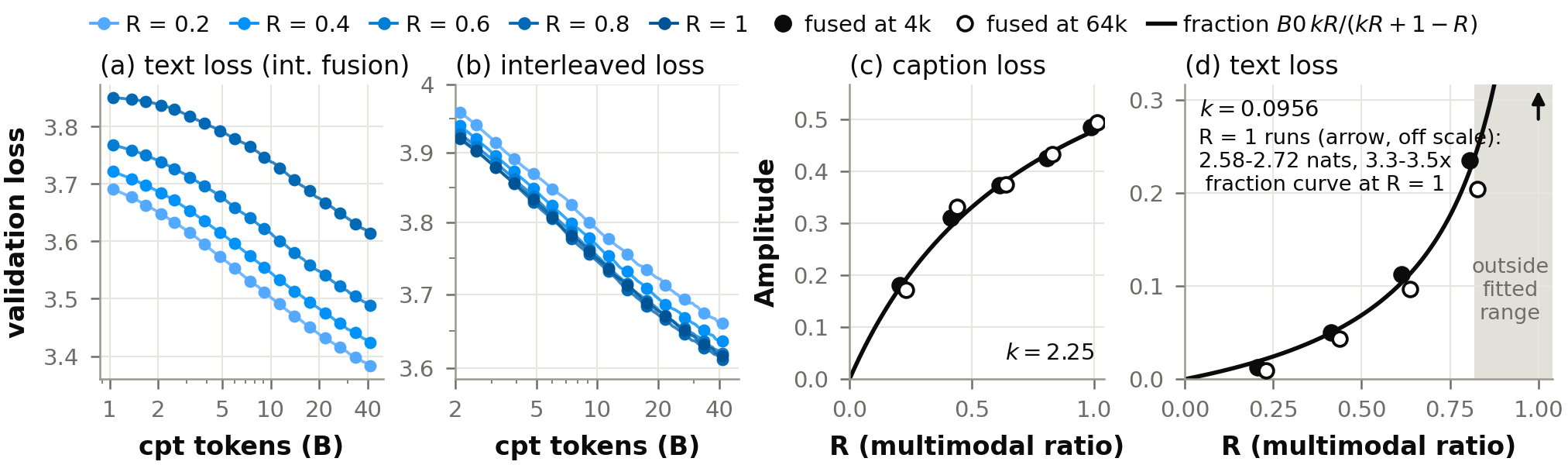}
\caption{\textbf{Mixture Ratio Dependency.}
\textbf{(a,b)} Text and interleaved validation loss curves for interleaved fusion varying the mixture ratio $R$.
\textbf{(c,d)} Fitted shift amplitude as a function of $R$ using the proposed fraction form for caption fusion, showing good fits at step 4,000 and robust validations at step 64,000.}
\label{fig:caption_curves}
\end{figure*}

We next consider more closely the distribution shift term $\Delta L$ and the multimodal fraction $R$.
We first perform interleaved fusion with various $R$ at a fusion point.
As expected, a smaller $R$ causes a smaller shift for text loss, and the opposite occurs for interleaved loss (\cref{fig:caption_curves}\textcolor{cvprblue}{a,b}).

\paragraph{Formulation.}
Furthermore, we see that $R$ curves are parallel to each other throughout $\Dcpt$. 
This indicates that there is no direct $R$-dependency on the base term (e.g., multiplicative $R$ factor would lead to curves with different $R$'s converging with $\Dcpt$).
The parallelism of the curves also suggests that the $R$-dependency can be factored out of the shift term: $ \Delta L = \sigma \cdot B_0 \cdot \rho(R) \cdot \Phi(\Dcpt)$,
where $\sigma \in \{+1, -1\}$ depending on the modality in consideration, $B_0$ a parameter to be fitted; $\Phi(\Dcpt)$ represents dependency on CPT steps, and $\rho(R)$ represents the $R$-dependent shift amplitude.
We propose to parameterize $\rho_m(R)$ as a \textit{fractional competition between modalities}:
\begin{equation}
\label{eq:rho_fraction}
\rho_m(R) = \frac{k_m R}{k_m R + (1 - R)},
\end{equation}
where $k_m$ represents the competitive pull of multimodal tokens relative to text.
 Since $\rho_m(0)=0$ and $\rho_m(1)=1$, $B_{0,m}$ directly represents the saturated shift amplitude.
 Furthermore, it is concave for $k>1$ and convex for $k<1$, one expression covers both curvatures (text loss is convex; multimodal loss is concave).

 In \cref{fig:caption_curves}\textcolor{cvprblue}{c,d}, we show that this form fits well to the $R$-dependent shift amplitude for both caption and text losses, except at the high $R$ regime for text loss where the text capabilites collapse. \footnote{In the same Fig. we further validate the fits at a different, larger value of the fusion point. 
 Furthermore, in Appendix \ref{app:more}, we compare our proposal with exponential forms \citep{ye2025data} showing that the fits are equally well and ours fits the high $R$ regime better.}
As CPT without text is undesirable in practice (destroying core text capabilities), we restrict the fitting domain of the caption-fusion $L_{\rm txt}$ scaling law to $R \le 0.82$, treating the collapse at $R = 1$ as the extreme boundary where the scaling law breaks down.

\paragraph{The $\Phi$ Term.}
We desire a functional form for $\Phi(\Dcpt)$ that grows from zero and saturates to one over CPT steps, capturing the buildup of the distribution shift (\cref{fig:curves}).
A simple form with this behavior would be $\Phi(D_{cpt}) = 1 - (1 + D_{cpt}/\tau)^{-\beta}$, with $\beta$,$\tau$ fitting parameters, $\tau$ being the saturation time. 
However, to elegantly relate it to the fraction term and $R$, we find a mathematical analogue in chemical kinetics \citep{laidler1987chemical}.

 In the (pseudo) reversible first-order reaction $A \underset{k_b}{\overset{k_f}{\rightleftharpoons}} B$, the equilibrium fraction $k_f/(k_f + k_b)$ corresponds to our shift amplitude $B_0 \rho(R)$, and the reaction rate provides a principled way to relate $\Phi(\Dcpt)$ to $R$.
 Borrowing this structure from chemistry, we propose $\Phi(\Dcpt)$ to be:
 \footnote{See Appendix \ref{app:reaction} for the chemical kinetics analogy details.}
\begin{equation}\label{eq:phi}
\Phi(\Dcpt) = 1 - \left(1 + (kR + 1 - R)\frac{\Dcpt}{\tau}\right)^{-\beta}
\end{equation}

\subsection{The Unified Scaling Law}
Assembling all the components, we have the final form of the unified scaling law:
\begin{equation}\label{eq:unified}
L = L_0 + A\, \mathcal{D}_{\text{base}}^{-\alpha} + \sigma\, B_0\,\rho(R)\, \Phi(\Dcpt)
\end{equation}
where $\sigma_{\text{cap}}=\sigma_{\rm int}=-1$, $\sigma_{\text{txt}}=+1$, and $\mathcal{D}_{\text{base}}$, $\rho$ and $\Phi$ are defined in Equations \ref{eq:dbase}, \ref{eq:rho_fraction} and \ref{eq:phi} respectively.
\cref{tab:params} shows the fitted parameters.

\begin{wrapfigure}{l}{0.5\textwidth}
\centering
\includegraphics[width=\linewidth]{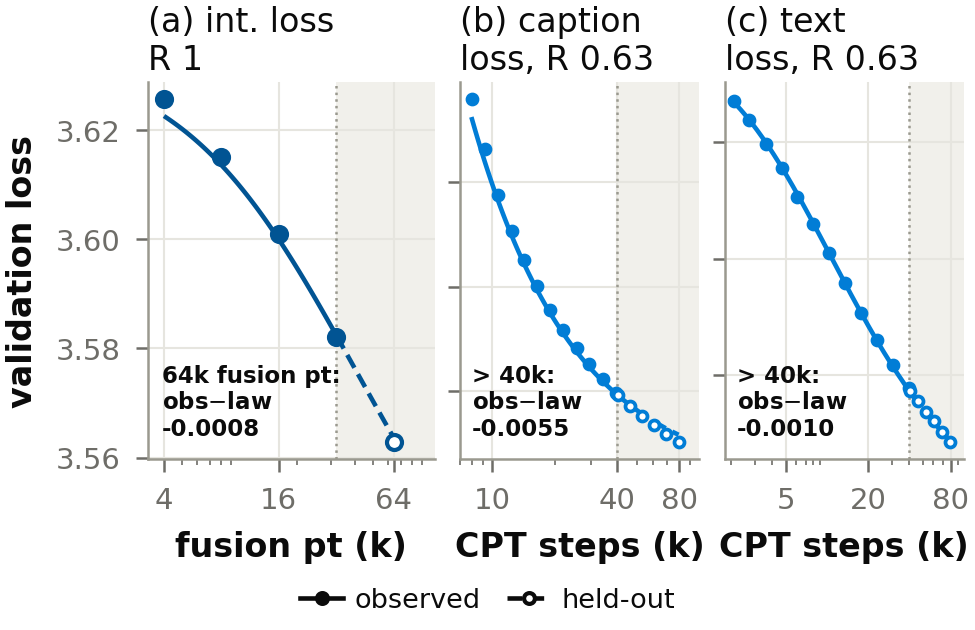}
\caption{\textbf{Validation of the scaling laws on held-out configurations beyond the fitting domain.} The law extrapolates to later fusion points and longer CPT horizons. More results in Appendix \ref{app:more}.}
\label{fig:extrapolation}
\end{wrapfigure}

\paragraph{Comparison with LLM CPT.}
Multimodal fusion exhibits a few empirical differences from unimodal text training: 
\begin{itemize}
    \item The parameter $w_m$ does not appear in standard LLM CPT laws \citep{wang2025learningdynamicscontinualpretraining}. 
    While our interleaved scaling law resembles LLM CPT when $w_{\rm int} \approx 1$, caption scaling exhibits a negligible $w_{\rm cap} \approx 0$, capturing a breakdown in cross-modal transfer under large distribution shifts.
    \item Training trajectories are inherently noisier than LLM CPT, and exhibit clearly an initial transient phase during early iterations (\cref{fig:curves}\textcolor{cvprblue}{d}). 
     We omit this transient period and focus on modeling the subsequent power-law regime, consistent with standard scaling law formulations \citep{kaplan2020scaling}.
\end{itemize}

\paragraph{Validation beyond the fit domain and Early Fusion.}
We further show that the scaling laws fit well on held-out validation data.
Validation performed includes fusion extrapolated to larger $\Dpt$ and extrapolation of CPT to larger $\Dcpt$ (\cref{fig:extrapolation}).
The overall RMSE is also shown in \cref{tab:params}.
Additionally, our scaling laws apply to early fusion, i.e., $\Dpt \to 0$.
See Appendix \ref{app:more} for more validation, extrapolation/interpolation and early fusion results.

\begin{findingbox}
    Both large and small distribution shift in multimodal fusion can be described with a unified scaling law.
    The base term unifies distinct modalities via cross-modal exchange rates, while the shift term is governed by a cross-modal competition in the form of a fraction.
\end{findingbox}

\section{Law Extensions}
\label{sec:law}

\subsection{Extension to Model Size}
\label{sec:size}
\begin{figure*}[t]
\centering
\includegraphics[width=\textwidth]{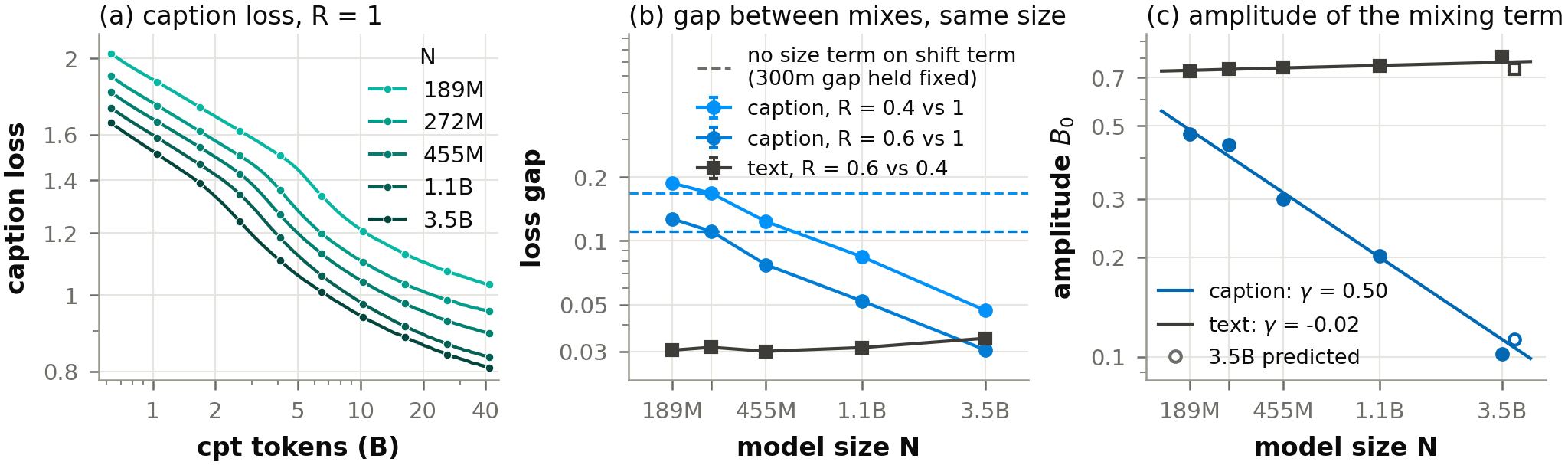}
\caption{\textbf{Scaling laws extended to varying model sizes ($N$).} \textbf{(a)} While caption loss decreases monotonically with scale , \textbf{(b)} the gap between different data mixtures ($L(R_1)-L(R_2)$) shrinks for captions but remains constant for text. \textbf{(c)} The amplitude $B_0$ of the shift term follows a power-law decay with model size for captions but is size-independent for text.}
\label{fig:size}
\end{figure*}

We next consider extending the law to include (total VLM) model size, $N$, where we previously held fixed.
Figure \ref{fig:size}\textcolor{cvprblue}{a} shows that the loss decreases monotonically with size as expected.

To describe this in the scaling law, the standard extension is incorporating a Chinchilla $N$ term \cite{hoffmann2022chinchilla} to the law: $C\, N^{-\eta}$. 
Under this extension, the loss gap between different $R$'s would be invariant to model size ($L(R_1,N)-L(R_2,N)= \text{constant w.r.t. } N$).

Empirical results directly refute this hypothesis (Figure~\ref{fig:size}\textcolor{cvprblue}{b}). 
On the caption axis, we see that the gap between $R=0.4$, $R=1$ and $R=0.6$, $R=1$ decreases with model size.
Conversely, on the text axis, the gap remains essentially unchanged.

We therefore parametrize the shift amplitude as a power law of model size, $B_0(N) = B_0\,N^{-\gamma}$, on top of the Chinchilla extension:
\begin{equation}\label{eq:size}
L = L_0  + A\,\mathcal{D}_{\rm base}^{-\alpha} + \sigma B_0\,\textcolor{red}{N^{-\gamma}}\, \rho(R)\,\Phi(\Dcpt) + \textcolor{red}{C\,N^{-\eta}}.
\end{equation}
Fitting this scaling law across model sizes yields $\gamma_{\rm cap}=0.5$, and we show that it is extrapolatable leaving out and validating the 3.5B endpoint (Figure~\ref{fig:size}\textcolor{cvprblue}{c}).
For text loss, we find $\gamma_{\rm txt}=0.02$, practically independent of model size.
We thus fix $\gamma_{\rm txt}=0$.

\begin{findingbox}
    The loss scales with an additive power-law term with respect to the model size. Furthermore, 
    the shift term decays with model size ($N^{-0.5}$).
\end{findingbox}

\subsection{Extension to Fusion with All Modalities}
\label{sec:threemodal}

\begin{wraptable}{l}{0.5\textwidth}
  \centering
  \footnotesize
  \setlength{\tabcolsep}{2pt}
  \begin{tabular}{cc cc cc}
  \toprule
  \multirow{2}{*}{$f$} & \multirow{2}{*}{\shortstack{Shares\\{\scriptsize($s_{\mathrm{t}} / s_{\mathrm{c}} / s_{\mathrm{i}}$)}}} & \multicolumn{2}{c}{Caption Res.} & \multicolumn{2}{c}{Text Res.} \\
  \cmidrule(lr){3-4} \cmidrule(lr){5-6}
   & & int $=$ txt & $w_{\mathrm{int}}{=}0.18$ & int $=$ cap & int $=$ txt \\
  \midrule
\multicolumn{6}{l}{\textit{Caption-free ($s_{\mathrm{cap}} = 0$)}} \\
  1.00 & .85 / .00 / .15 & -- & -- & $-0.012$ & $+0.007$ \\
  1.00 & .40 / .00 / .60 & -- & -- & $-0.077$ & $+0.048$ \\
  \midrule
  \multicolumn{6}{l}{\textit{Low interleaved ($f \le 0.25$): baseline law holds}} \\
  0.22 & .38 / .48 / .14 & $-0.016$ & $-0.011$ & $-0.025$ & $+0.023$ \\
  \midrule
  \multicolumn{6}{l}{\textit{High interleaved ($f > 0.25$): exchange rate $w_{\mathrm{int}}$ needed}} \\
  0.33 & .40 / .40 / .20 & $-0.029$ & $-0.020$ & $-0.034$ & $+0.027$ \\
  0.75 & .20 / .20 / .60 & $-0.057$ & $+0.014$ & $-0.140$ & $+0.082$ \\
  \bottomrule
\end{tabular}
  \caption{\textbf{Evaluation residuals (nats) for three-modality mixtures across interleaved share $f$ (abbreviated).} Caption loss evaluating interleaved tokens as text versus exchange rate $w_{\mathrm{int}}{=}0.18$. Text loss evaluating them as caption and text bounds the true loss. Full results in Appendix \ref{app:more}.}
  \label{tab:threemodal2}
\end{wraptable}

We extend the study to training on text, caption and interleaved data at once.
We consider a fixed caption/interleaved blend, with the interleaved fraction of the multimodal side denoted by $f$ (so the ratios are $s_{\mathrm{int}} = f\,R$, $s_{\mathrm{cap}} = (1 - f)R$ and $s_{\mathrm{txt}} = 1 - R$.), treating it only as a small perturbation to the caption fusion scaling law, for practical reasons discussed in the beginning of \cref{sec:interleaved} (interleaved data do not infuse strong visual capabilities compared to caption data).

The simplest extension is treating interleaved tokens as part of text tokens; as we have shown in \cref{sec:interleaved}, the distribution shift is mild compared to caption data.
Simply counting the interleaved token as text, we find that the scaling law for caption loss holds for $f \lesssim 0.25$ (\cref{tab:threemodal}).
For text loss, treating the interleaved token as caption and text respectively upper and lower bound the observed values.

The prediction of caption law can be slightly improved with the following method.
Let $w_{\mathrm{int}}$ be the exchange rate of an interleaved token with respect to caption.
We can reuse our two-source scaling law by assigning interleaved tokens an exchange rate $w_{\mathrm{int}}$ such that $\mathcal{D}_{\mathrm{base,cap}}$ is modified to:
\begin{equation}\label{eq:slice}
w_{\mathrm{cap}}\,\Dpt + \bigl[(1 - f) + f\,w_{\mathrm{int}}\bigr]\,R\,\Dcpt + w_{\mathrm{cap}}\,(1 - R)\,\Dcpt , \nonumber
\end{equation}
which adds a single parameter $w_{\mathrm{int}}$ to the caption law.
Fitting yields $w_{\mathrm{int}} \approx 0.18$, meaning that an interleaved token is therefore worth about a fifth of a caption token for the caption loss.
This yields better fitting at low and high $f$'s (\cref{tab:threemodal2}).

\begin{findingbox}
    The caption scaling law extends to three-modality mixtures by treating interleaved data as text at low shares ($f \le 0.25$), or by assigning it an effective exchange rate ($w_{\text{int}} = 0.18$) as the interleaved fraction scales higher ($f > 0.25$).
    \end{findingbox}

\section{Early/Late Fusion Trade-offs}
\label{sec:timing}

\subsection{Base VLM Evaluation}
\label{sec:downstream_eval}

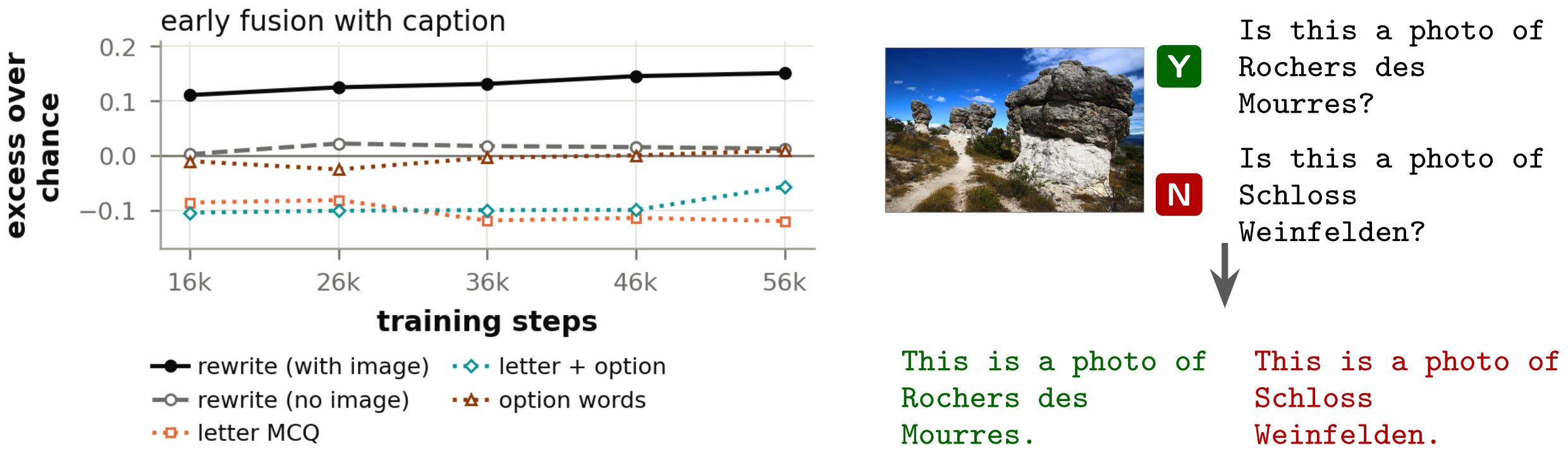
\begin{figure*}[t]
\centering
\includegraphics[width=0.52\textwidth]{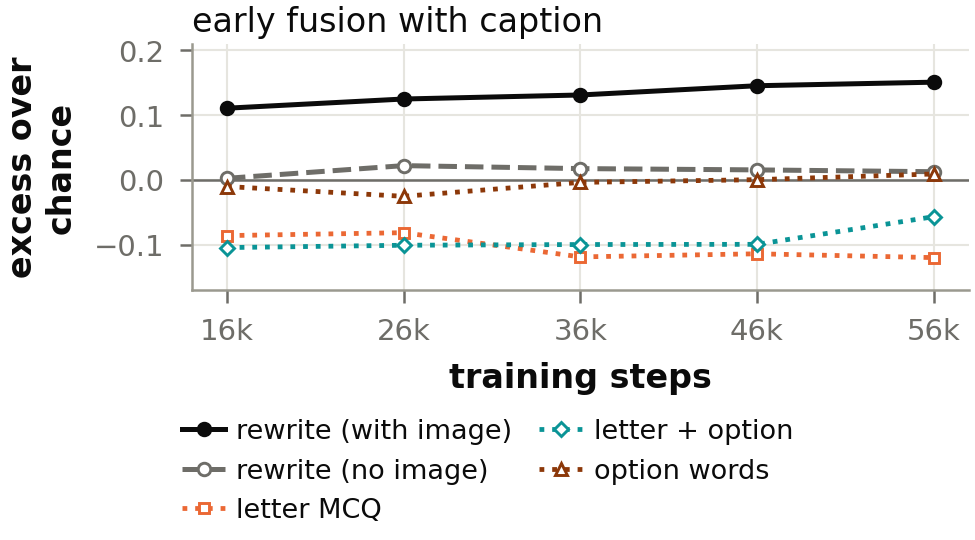}\hfill
\resizebox{!}{0.288\textwidth}{%
\begin{tikzpicture}[
  qline/.style={font=\ttfamily\scriptsize, align=left, anchor=west},
  badge/.style={rounded corners=1.5pt, inner sep=2.2pt,
                font=\sffamily\scriptsize\bfseries, text=white},
]

\node[draw=black!50, inner sep=0pt, anchor=south west] at (0,0.38)
  {\includegraphics[height=1.25cm]{mme_pair_00488.jpg}};

\node[badge, fill=green!40!black] at (2.25,1.50) {Y};
\node[qline] at (2.58,1.50) {Is this a photo of\\ Rochers des\\ Mourres?};
\node[badge, fill=red!70!black] at (2.25,0.52) {N};
\node[qline] at (2.58,0.52) {Is this a photo of\\ Schloss\\ Weinfelden?};

\draw[-{Stealth[length=3mm]}, line width=1.4pt, draw=black!65]
  (2.60,0.15) -- (2.60,-0.35);

\node[qline, text=green!40!black, anchor=north west] at (0,-0.55)
  {This is a photo of\\ Rochers des\\ Mourres.};
\node[qline, text=red!70!black, anchor=north west] at (2.70,-0.55)
  {This is a photo of\\ Schloss\\ Weinfelden.};

\end{tikzpicture}%
}
\caption{\textbf{(left) MME benchmark accuracies over training for from-scratch multimodal (caption) pre-training of 1b VLM.}
We compare our rewriting strategy (with and without image prefix), with naive scoring with continuation formats (letter, letter + option, option words).
Our strategy maintains a positive visual grounding gap (scores with image prefix higher than without) and steady learning progression, while naive formats collapse to chance or produce erratic scores due to prior biases. More results in Appendix \ref{app:eval}.
\textbf{(right) Illustration of the rewriting strategy used for downstream evaluation of base VLMs.} Multiple-choice questions are converted into statements conditioned on the image prefix to enable log-likelihood ranking and avoid answer-prior collapse.}
\label{fig:fig_same_item_formats}
\label{fig:rewrite}
\end{figure*}

In addition to validation loss, we seek to evaluate the multimodal pre-trained models via downstream vision benchmarks. 
In text-only language modeling, pre-trained only base checkpoints are evaluated without requiring instruction-following (IF) abilities by measuring the cloze-style (continuation) reference answer's perplexity or ranking candidate sentences (LAMBADA \citep{paperno-etal-2016-lambada}, HellaSwag \citep{hellaswag}). 

In multimodal learning, however, standard visual question answering (VQA) and reasoning benchmarks (MME~\cite{fu2026mme}, GQA~\cite{hudson2019gqa}, MMMU~\cite{yue2024mmmu}) are predominantly designed for IF models.
  Base VLMs cannot be evaluated directly on them, and supervised fine-tuning (SFT) is not an appropriate remedy either, as it introduces substantial confounders (SFT data mixture, format bias, new vision-text alignment, etc.) that obscure the pre-training fusion dynamics under study.

To evaluate base models without IF, an intuitive alternative is naive text continuation perplexity (evaluating the reference answer's perplexity given an image and question prompt).
This works partially for DocVQA~\cite{mathew2021docvqa} but fails on multiple-choice and binary tasks (MME, GQA); 
the model's output is dominated by drifting answer priors (e.g., alternating between predicting exclusively ``Yes'' and ``No'', or collapsing onto a single option letter) rather than visual understanding, yielding ungradable outputs or chance-level accuracy (\cref{fig:fig_same_item_formats}).

\paragraph{Our Rewriting Strategy.}
To address this, we map each question-candidate pair into a declarative assertion conditioned on the image prefix (\cref{fig:rewrite}), and score them via byte-normalized log-likelihood ranking.
This removes format-following demands while providing sufficient context to eliminate answer-prior collapse.
We find that under this protocol, models evaluated on MME and GQA demonstrate visual learning above-chance scoring along the training trajectory while maintaining positive visual grounding gap (\cref{fig:fig_same_item_formats}). 
See Appendix \ref{app:eval} for details of the rewriting strategy and other results.

\begin{findingbox}
  Converting multiple-choice benchmarks to log-likelihood ranking benchmarks helps avoid answer-prior collapse and extract the visual capabilities of a base VLM without requiring instruction-following tuning.
\end{findingbox}

\subsection{Optimizing Fusion Timing}
\label{sec:fusion}
Armed with scaling laws and tools developed thus far, we here analyze the trade-offs across different fusiong timings.
Specifically, we are interested in how the fusion timing $\Dpt$ affects the performance of the model under a fixed budget of text and caption tokens ($T_{\mathrm{txt}}$, $T_{\mathrm{cap}}$ respectively).
Formally, given the budget, we want to find the optimal fusion timing parameterized by the text-only pre-training duration $D_{\text{pt}} \in [0, T_{\text{txt}}]$ that minimizes the caption loss $L_{\text{cap}}$ and/or text loss $L_{\text{txt}}$:
$$\argmin_{\Dpt\in [0, T_{\text{txt}}]} L_{\text{cap}}(D_{\text{pt}}) \quad \text{and/or} \quad \argmin_{D_{\text{pt}} \in [0, T_{\text{txt}}]} L_{\text{txt}}(D_{\text{pt}})$$
Under fixed token budgets $(T_{\text{txt}}, T_{\text{cap}})$, the base term simplifies to $\mathcal{D}_{\text{base,cap}} = w_{\rm cap} T_{\text{txt}} + T_{\text{cap}}$ (and analogously for text), remaining constant regardless of fusion timing $D_{\text{pt}}$.
Therefore, the trade-off is determined solely by the shift term.

As proven in Appendix~\ref{app:fusion}, the caption shift magnitude $|\Delta L_{\text{cap}}|$ (and analogously for text) is strictly monotonically increasing with $D_{\text{pt}}$, pushing optimal solutions to the edge:
\begin{align}
&\argmin_{\Dpt} L_{\text{cap}}  = T_{\rm txt}\implies D_{\text{cpt}}^* = T_{\text{cap}}, \;   R^* = 1  \\
&\argmin_{\Dpt} L_{\text{txt}} = 0\implies D_{\text{cpt}}^* = T_{\text{txt}} + T_{\text{cap}}, \, R^* = \frac{T_{\text{cap}}}{T_{\text{txt}} + T_{\text{cap}}}
\end{align}
In other words, late fusion ($\Dcpt$ shortest and $R$ densest) favors lower caption loss while early fusion ($\Dcpt$ longest and $R$ sparsest) favors lower text loss.

We further note that $B_{0,\text{cap}}(N)$ decays with model size (\cref{sec:size}). 
This means that the late fusion advantage, captured by the shift term, diminishes with larger model size, and early fusion becomes more favorable for larger models.

\begin{table*}[th]
\centering
\footnotesize
\setlength{\tabcolsep}{5.5pt}
\begin{tabular}{@{}llcc cc ccc ccc@{}}
\toprule
& & & & \multicolumn{2}{c}{\textbf{Val Loss (nats)}} & \multicolumn{2}{c}{\textbf{Text Benchmarks}} & \multicolumn{3}{c}{\textbf{Vision Benchmarks}} \\
\cmidrule(lr){5-6}\cmidrule(lr){7-8}\cmidrule(lr){9-11}
\textbf{Scale} & \textbf{Schedule} & $\Dpt$ & $R$ & \textbf{Caption} $\downarrow$ & \textbf{Text} $\downarrow$ & \textbf{HSwag} $\uparrow$ & \textbf{LAMB} $\downarrow$ & \textbf{MME} $\uparrow$ & \textbf{GQA} $\uparrow$ & \textbf{DocVQA} $\downarrow$ \\
\midrule
\multirow{5}{*}{\textbf{300M}} 
& S0 (Early) & 0   & 0.45 & 1.1488 & 3.4095 & 0.3691 & 69.8   & 0.6701 & 0.4103 & 3.88 \\
& S1         & 4k  & 0.48 & 1.1418 & 3.4194 & 0.3668 & 73.0   & 0.6796 & 0.3694 & 3.73 \\
& S3         & 16k & 0.63 & 1.0983 & 3.4574 & 0.3613 & 89.3   & 0.6779 & 0.3891 & 4.13 \\
& S4         & 24k & 0.78 & 1.0486 & 3.5341 & 0.3591 & 80.0   & 0.6883 & 0.3969 & 3.78 \\
& S5 (Late)  & 32k & 1.00 & 0.9902 & 6.3135 & 0.2938 & 2466.4 & 0.6770 & 0.3869 & 3.95 \\
\cmidrule(lr){2-11}
& \multicolumn{3}{l}{\textit{Spearman $r_s$}} & -- & -- & $-0.94$ & $+0.94$ & $-0.20$ & $+0.37$ & $-0.31$ \\
\midrule
\multirow{5}{*}{\textbf{1B}} 
& S0 (Early) & 0   & 0.45 & 0.9659 & 3.0258 & 0.4862 & 27.0   & 0.7064 & 0.4134 & 3.10 \\
& S1         & 4k  & 0.48 & 0.9600 & 3.0360 & 0.4795 & 27.3   & 0.7055 & 0.4236 & 2.91 \\
& S3         & 16k & 0.63 & 0.9304 & 3.0791 & 0.4702 & 24.8   & 0.7116 & 0.4158 & 2.94 \\
& S4         & 24k & 0.78 & 0.9029 & 3.1589 & 0.4569 & 32.3   & 0.7090 & 0.4159 & 2.67 \\
& S5 (Late)  & 32k & 1.00 & 0.8705 & 5.8537 & 0.3144 & 557.6  & 0.6839 & 0.4209 & 2.80 \\
\cmidrule(lr){2-11}
& \multicolumn{3}{l}{\textit{Spearman $r_s$}} & -- & -- & $-1.00$ & $+0.70$ & $+0.20$ & $-0.40$ & $+0.80$ \\
\bottomrule
\end{tabular}
\caption{\textbf{Validation losses and downstream benchmark performance across fusion timings (abbreviated).} 
Caption loss decreases with delayed fusion, while text loss increases.
Text benchmarks closely track validation text loss (collapsing at S5, $R=1.0$), whereas downstream vision benchmarks fail to track validation caption loss. Full results in Appendix \ref{app:eval}.}
\label{tab:downstream_merged}
\end{table*}

\paragraph{Empirical Evidence.}
We run fusion timing experiments to test if the above conclusion holds in practice.
We fix the total budget to 56B tokens, and sweep the fusion timing, including early fusion (labeled S0 to S5), across 300M and 1B models.
Besides measuring the validation losses, we evaluate the models' downstream performance using methods proposed in \cref{sec:downstream_eval}.

\cref{tab:downstream_merged} shows the main results.
Validation loss behaviors conform to our predictions, as caption (text) loss decreases (increases) with delayed fusion timing.
Moreover, we notice that the caption loss gap between early and late fusion (S0-S5) narrows from 0.159 to 0.095 (40\% reduction), while the text loss gap changes mildly, from 2.904 to 2.828 (2.6\% relative change), which are in accordance to our law prediction.
Notably, the vanishing absolute loss gap also translates to a vanishing \textit{token-equivalent} gap: since the base learning trajectory (main driver of loss decrease via extra tokens) is independent of model size (\cref{eq:unified}), the extra caption token budget required by early fusion to catch up with late fusion's vision performance drops with model size from $\sim 39$B tokens (300M model) to $\sim 23$B (1B model) in our example.
On the other hand, as the text shift amplitude is independent of model size, the number of tokens required by late fusion to match early fusion's text loss remains constant at any scale.

For downstream evaluation, results show a notable divergence between text and vision benchmarks. 
On text benchmarks, downstream capabilities closely reflect validation text loss. 
Across both 300M and 1B scales, accuracy benchmarks correlate strongly and negatively with text loss (see the Spearman's rank correlation coefficients in \cref{tab:downstream_merged}).
The LAMBADA perplexity correlates positively with text loss, and notably explodes by orders of magnitude at S5 ($R=1$), in line with our characterization of the $R=1$ collapse regime in Section \ref{sec:interleaved}.

On the other hand, downstream performance of vision benchmarks fails to track validation caption loss. 
MME has little correlation with caption loss, and all three benchmarks' correlations are inconsistent across scale.
We do observe that benchmark scores improve when both text and caption losses decrease simultaneously across scale, indicating the loss is still a valid performance proxy only when both modalities are jointly improved. 
One possible explanation for this is that VQA requires both language and vision capabilities, and optimizing one of them while degrading the other may not be optimal for downstream performance.
Taken together, these findings strongly reinforce early fusion as the preferred recipe as late fusion penalizes language capabilities without delivering tangible downstream vision gains.

\begin{findingbox}
  Studying fusion timing under a fixed budget reduces to analyzing the shift term, where in theory and in practice, early fusion minimizes text loss while late fusion minimizes vision loss, with the vision advantage of late fusion fading with model size.
  Since text benchmarks mirror the text loss behavior while vision benchmark scores do not reflect caption loss improvement, early fusion is overall the preferred pre-training recipe. 
\end{findingbox}

\section{Conclusion}
\label{sec:conclusion}

In this paper, we have presented scaling laws for vision-language fusion, by decomposing the learning dynamics into a base modality progress and an additive competitive shift, unifying both low-shift (interleaved) and high-shift (caption) regimes.
The scaling laws also reveal that early fusion minimizes text loss while late fusion minimizes vision loss.
However, as the late-fusion vision gains decay with model size and fails to translate to downstream gains, early fusion proves superior in practice.

\paragraph{Limitations and Future Work.}
As in standard scaling law literature, the exact fitted parameters depend on design choices (data, tokenization, etc.) that may not transfer; thus, our formulation may better be seen as a framework for practitioners to characterize and fit on their own training rather than rigid constants.
Additionally, our formulation omits regimes not viable in practice, i.e., extreme caption ratios ($R \to 1$) that cause language collapse (\cref{sec:shift_and_competition}). 
Finally, scaling to larger models and broader modalities remains exciting future work.

\section*{Acknowledgement}
This research was conducted by Noetra Corp. under FRONTia, a Japanese national program led by the Ministry of Economy, Trade and Industry of Japan (METI) and the New Energy and Industrial Technology Development Organization (NEDO), aimed at developing a domestic multimodal foundation model for AI robots and physical AI.

\bibliographystyle{plainnat}
\bibliography{main}

\clearpage

\appendix

\section{Setup Details}
\label{app:setup}

\begin{wraptable}{l}{0.5\textwidth}
\centering
\small
\setlength{\tabcolsep}{3.5pt}
\begin{tabular}{lccccc}
\toprule
\textbf{Label} & \textbf{Dim} & \textbf{FFN} & \textbf{LLM} & \textbf{Non-Emb} & \textbf{VLM} \\
\midrule
200M  & 400  & 1280  & 92.8M   & 49.8M   & 189.3M  \\
300M & 640  & 2048  & 175.0M  & 106.2M  & 272.2M  \\
500M & 1040 & 3328  & 356.3M  & 244.5M  & 454.7M  \\
1B   & 2000 & 6400  & 1017.0M & 801.9M  & 1118.4M \\
3.5B & 4000 & 12800 & 3416.4M & 2986.3M & 3523.9M \\
\bottomrule
\end{tabular}
\caption{\textbf{Model architecture across scales used in the paper.} Columns give LLM parameters, non-embedding parameters, and total VLM parameters (including the vision encoder and projector).}
\label{tab:architecture}
\end{wraptable}

We describe in more detail the training setup of our study.

\paragraph{Model.}
We adopt Gemma-3 as the base architecture for our study.
The number of layers is fixed to 18; head dimension is 256 with 4 query and 2 KV heads.
We scale the model size via its feedforward neural network (FFN) layer width (Figure \ref{tab:architecture}).
The embedding is tied, and the vocabulary size is 107,520.
The SigLIP2 ViT-B encoder in use converts each $16\times16$ patch to an embedding.

\paragraph{Data.}
For text-only pre-training, we use the original FineWeb-Edu corpus without modification.
For interleaved data (OBELICS and OmniCorpus), we apply text-quality filtering following the FineWeb-Edu classification heuristics to match FineWeb-Edu corpus quality.
For image-caption data (ReLAION), we rewrite the alt-text annotations to produce higher-quality captions.

\paragraph{Hyperparameters and Training Details.}
We summarize the hyperparameters used for training and other training details in Table \ref{tab:hyperparams}.

We train with mixed precision and FSDP using a codebase based on \texttt{torchtitan} \cite{liang2025torchtitan}.
Additionally, sequence packing is performed to retain token processing efficiency.

During CPT, all components' neural network parameters are unfrozen.
We perform warmup only in the beginning of PT and use a constant learning rate throughout the remaining PT and CPT phases.
See Appendix \ref{app:anneal} for discussion of learning rate annealing.

\paragraph{Fitting.}
We mainly follow the procedure described in \cite{besiroglu2024chinchilla} to perform fitting.
Particularly, the logarithm of loss is fitted with Huber loss with $\delta=10^{-3}$.
L-BFGS-B is used to optimize the objective.
Instead of grid search, we initialize the coefficients randomly a few hundred times and run L-BFGS-B to optimize the objective.

\section{More Results}
\label{app:more}
\subsection{More on Scaling Law Formulation}
\begin{figure*}[t]
\centering
\includegraphics[width=\textwidth]{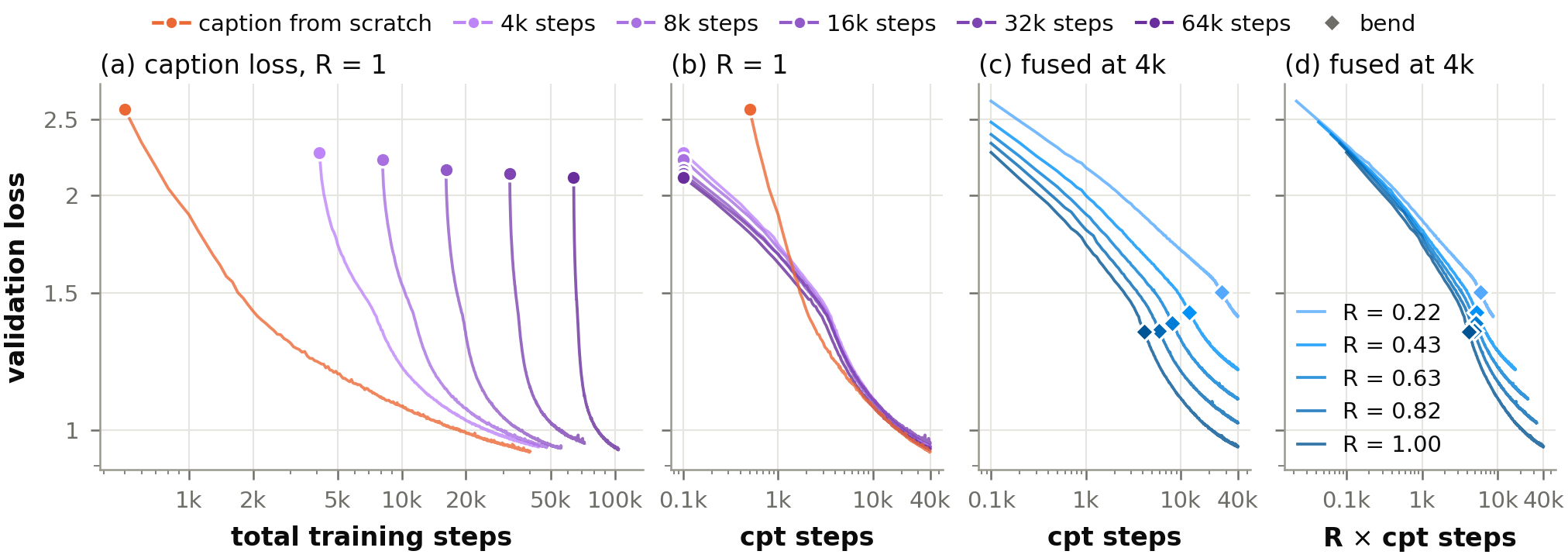}
\caption{\textbf{Caption validation loss dynamics across training stages.} The two panels on the left show that text pre-training provides negligible benefit to caption loss. The two panels on the right show that the loss gap across varying mixture ratios ($R$) remains almost constant when measured by effective caption tokens ($R \times$ CPT steps) rather than raw CPT steps.}
\label{fig:fig_caption}
\end{figure*}

\paragraph{More on the Cross-Modal Exchange Rate.}\par
\begin{wraptable}{l}{0.57\textwidth}
\centering
\footnotesize
\begin{tabular}{ll}
\toprule
\textbf{Hyperparameter} & \textbf{Value} \\
\midrule
Peak Learning Rate & $4 \times 10^{-4}$ \\
Learning Rate Schedule (PT) & 500-step linear warmup, then flat \\
Learning Rate Schedule (CPT) & Flat from step 1 (no warmup) \\
Batch Size & $\approx 1.05$M tokens (256 seq $\times$ 4096) \\
Context Length & 4,096 tokens \\
Optimizer & AdamW \\
Adam $\beta_1, \beta_2$ & 0.9, 0.95 \\
Adam $\epsilon$ & $10^{-8}$ \\
Weight Decay & 0.1 \\
Gradient Clipping Norm & 1.0 \\
\bottomrule
\end{tabular}
\caption{\textbf{Training hyperparameters for the language (PT) and vision-language continued pre-training (CPT) pipelines.} These values remain fixed across all evaluated model sizes.}
\label{tab:hyperparams}
\end{wraptable}

We give more justification for the use of $R \Dcpt$ as the effective caption token count in the scaling law formulation.
Note that the loss curve in \cref{fig:fig_caption}\textcolor{cvprblue}{b} (same as \cref{fig:curves}\textcolor{cvprblue}{c}, repeated for convenience) has a bend transitioning to the power-law regime, where the the law starts describing the learning.
If text tokens after fusion advanced learning as much as caption tokens
do, runs with different $R$ would reach this bend after a similar number
of CPT steps.

\begin{figure*}[t]
\centering
\includegraphics[width=\textwidth]{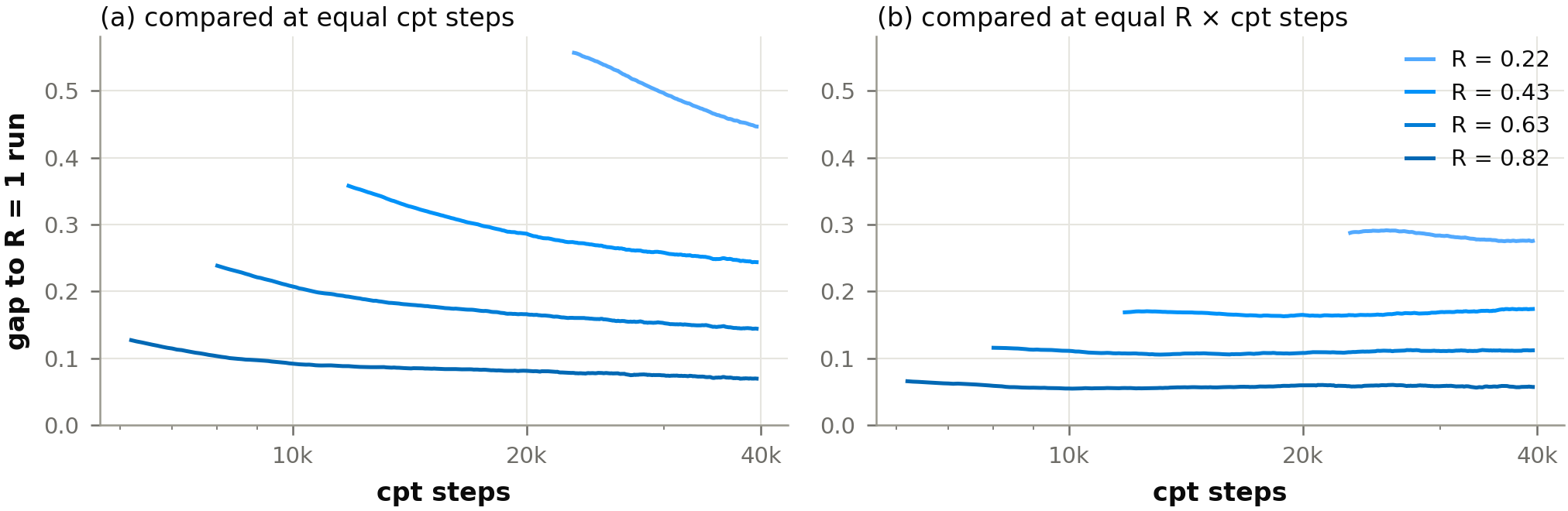}
\caption{\textbf{The validation loss gap between mixed runs ($R < 1$) and the multimodal-only run ($R = 1$).} The gap shrinks when compared at equal CPT steps (left) but remains nearly constant when aligned by effective caption tokens ($R \times$ CPT steps) (right).}
\label{fig:caption_gap_curves}
\end{figure*}

However, as shown in \cref{fig:fig_caption}\textcolor{cvprblue}{c}, the bend spreads from around 4,000 steps to 30,000 steps. 
When we count the steps in caption tokens, $R\,\Dcpt$, as in the fourth panel in \cref{fig:fig_caption}\textcolor{cvprblue}{d}, the spread is much smaller (4,000 to 6,000 caption-only steps).
Furthermore, the gap of the loss curves among different $R$'s stay almost constant at equal $R\,\Dcpt$, whereas the gap shrinks at equal CPT steps (see \cref{fig:caption_gap_curves}  for comparison). 
The former observation is consistent with what we observe with the interleaved data in Section \ref{sec:interleaved}.

\paragraph{More on the Shift Term.}
The $R$-dependent $\rho(R)$ can be parameterized with exponential functions, following previous studies on data mixtures \cite{ye2025data,khandelwal2026scaling}:
\begin{equation}\label{eq:int_f}
\rho_{\rm txt}(R)=  e^{a_{\text{txt}} R} - 1;\quad \rho_{\rm int}(R)= 1 - e^{-a_{\text{int}}R}
\end{equation}
with $a_{\text{txt}}$ and $a_{\text{int}}$ as fitting parameters.
Note that the functional form is designed such that the law falls back to the base term without shift when $R=0$.

\begin{figure*}[t]
\centering
\includegraphics[width=\textwidth]{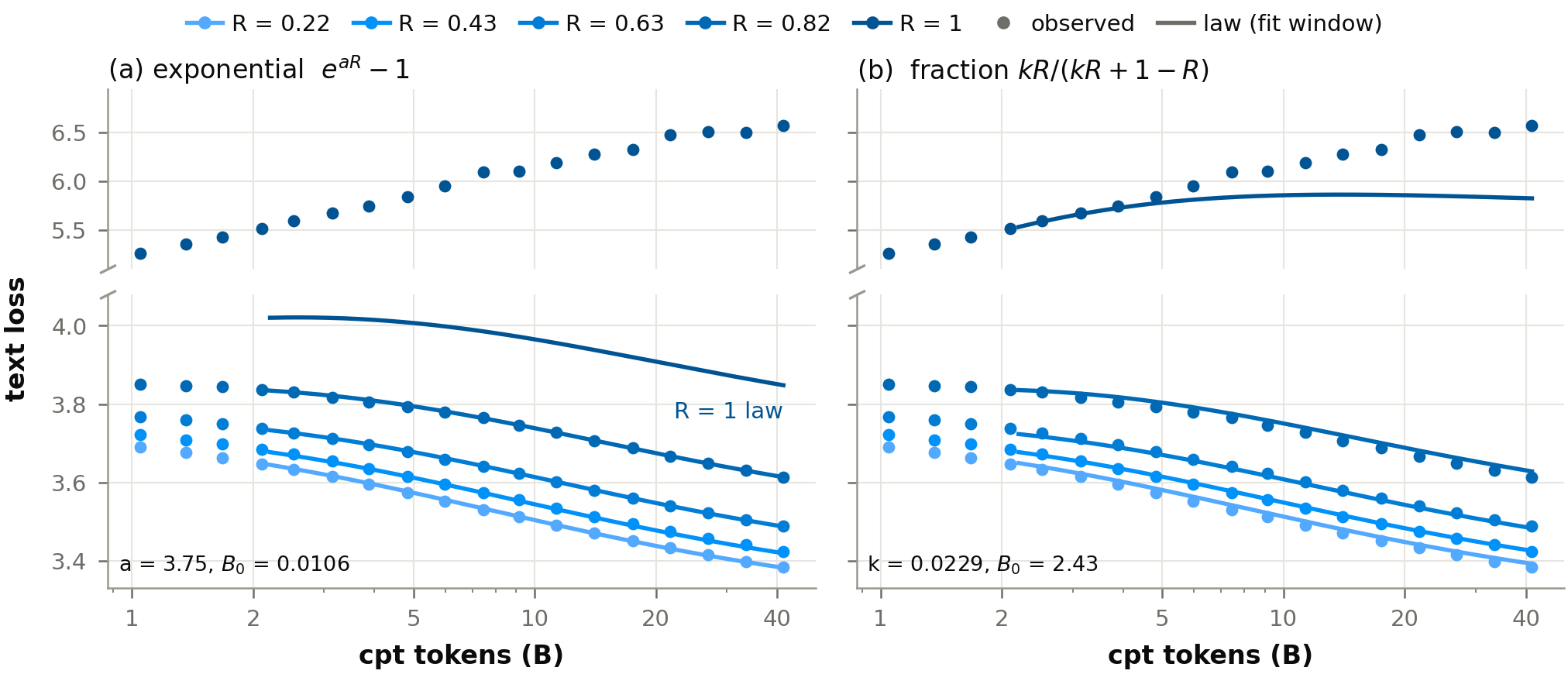}
\caption{\textbf{Comparison of the exponential and fraction formulations for modeling the mixture ratio ($R$) dependency of the text loss shift term.} The fraction form $kR/(kR+1-R)$ better tracks both the mixed ($R<1$) and unmixed ($R=1$) training regimes.}
\label{fig:text_ramp}
\end{figure*}

Besides being more compact, our proposed fraction form also fits at least as well
and generalizes better across data mixture regimes. 
The difference is most prominent at the edge of data mixture.
As in Figure \ref{fig:text_ramp}, exponential parameterization cannot fit simultaneously the mixed ($R < 1$) and unmixed runs ($R = 1$).
In contrast, the fraction form tracks both regimes with better fidelity.

\begin{figure*}[t]
\centering
\includegraphics[width=\textwidth]{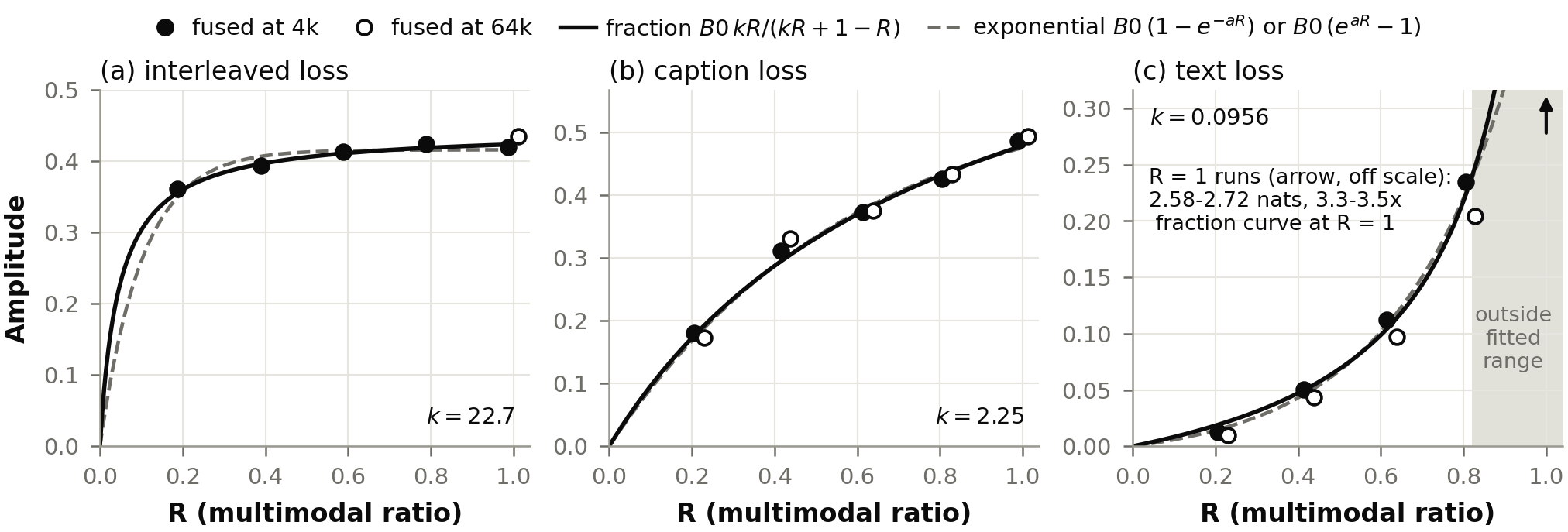}
\caption{\textbf{Fitted shift amplitude as a function of the multimodal mixture ratio ($R$) using exponential and fraction forms.} The single fraction form unifiedly captures the concave curvature for interleaved/caption losses and the convex curvature for text loss using a single parameter $k_m$. }
\label{fig:ramp}
\end{figure*}

Nevertheless, incorporating the $R = 1$ collapse regime noticeably inflates residual error across the interior mixed runs.
As CPT without text is undesirable in practice (destroying core text capabilities), we restrict the fitting domain of the text scaling law to $R \le 0.82$, treating the collapse at $R = 1$ as the extreme boundary where the scaling law breaks down.

Within this range, the fraction form matches or outperforms the exponential form at describing the $R$ dependency, while unifying all three evaluation axes under a single parameter $k = 22.7$, $2.25$ and $0.0956$ for interleaved, caption and text loss respectively (\cref{fig:ramp}).

\subsection{Validation Results, Early Fusion, and Three-Modality Mixtures}
\begin{figure*}[t]
\centering
\includegraphics[width=\textwidth]{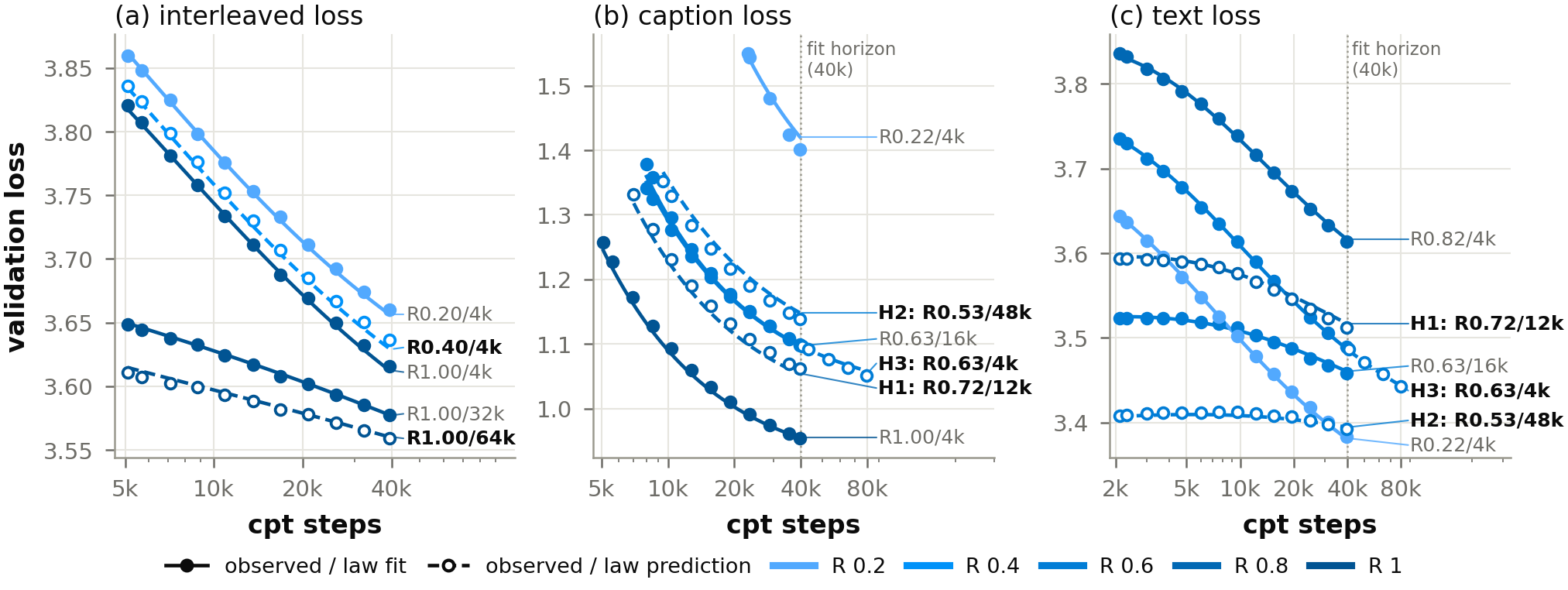}
\caption{\textbf{Validation loss trajectories for both fitted runs and held-out configurations (extrapolated fusion points and CPT horizons) across interleaved, caption, and text modalities.}}
\label{fig:fig_validation}
\end{figure*}

\begin{figure*}[t]
\centering
\includegraphics[width=\textwidth]{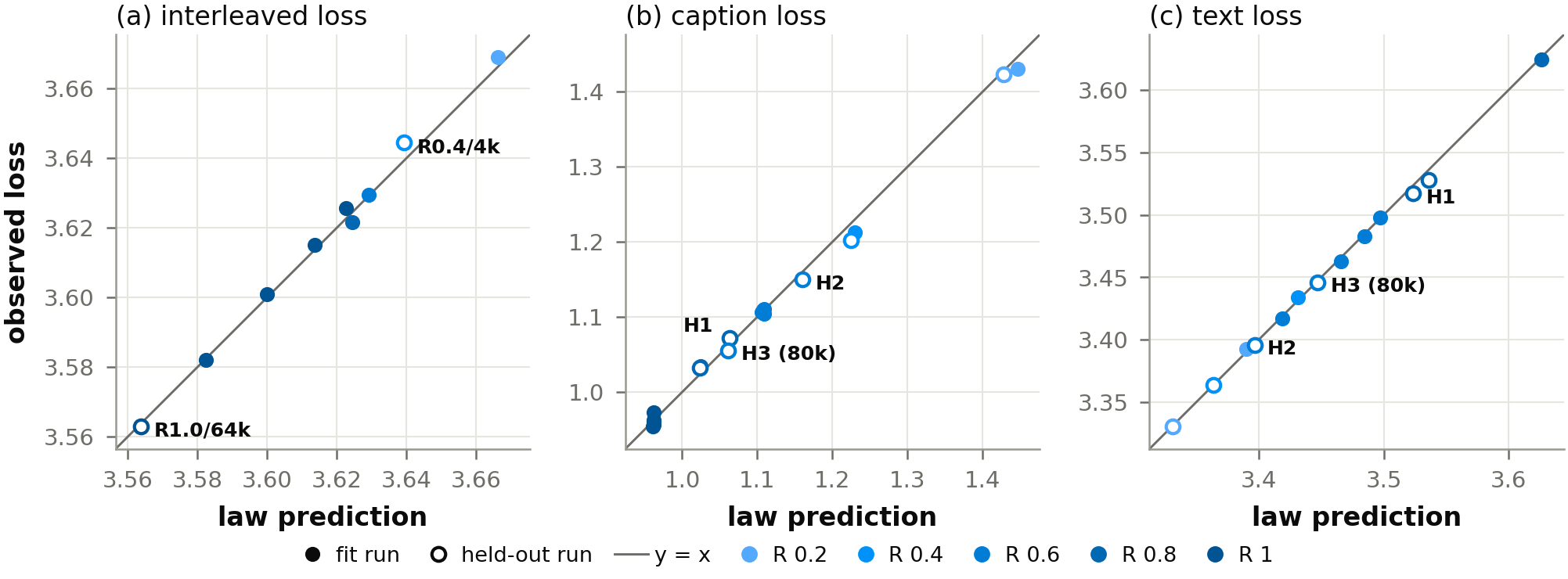}
\caption{\textbf{Observed versus predicted validation losses for fitted and held-out validation runs.} The law predictions tightly align with observed losses across varying configurations.}
\label{fig:fig_validation_parity}
\end{figure*}

We run held-out validation to test our scaling laws rigorously, varying $\Dpt,R,\Dcpt$.
The runs are:
\begin{itemize}
    \item Interleaved held out evaluation \texttt{R0.40/4k}: validating interpolation at $R=0.4$
    \item Interleaved held out evaluation \texttt{R1.00/64k}: validating extrapolation at $\Dpt=$ 64,000 steps.
    \item Text-caption held out evaluation \texttt{H1}: joint fusion and ratio interpolation at $\Dpt=$ 12,000 steps and $R=0.722$
    \item Text-caption held out evaluation \texttt{H2}: late-fusion interpolation at $\Dpt=$ 48,000 steps with $R=0.527$
    \item Text-caption held out evaluation \texttt{H3}: CPT extrapolation at $\Dpt=$ 4,000 steps with $R=0.63$ and $\Dcpt=$ 80,000 steps.
\end{itemize}
The results are shown in \cref{fig:fig_validation,fig:fig_validation_parity}.
The held-out RMSE's reported in Table \ref{tab:params} are the average over these validation runs.

\begin{figure*}[t]
\centering
\includegraphics[width=\textwidth]{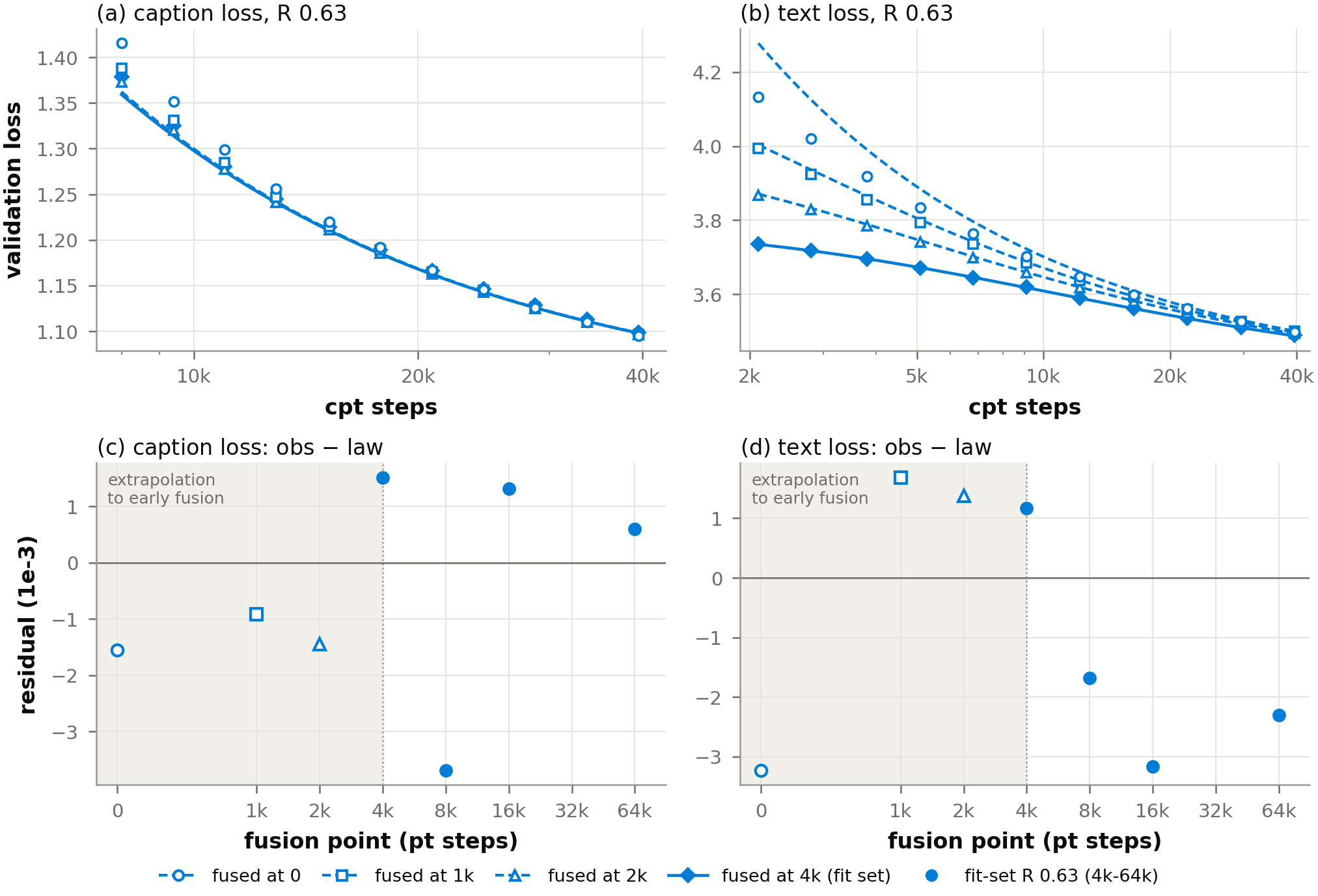}
\caption{\textbf{Extrapolation of the scaling law to early fusion regimes ($D_{pt} = 0$).} The top panels map the predicted versus observed loss trajectories, while the bottom panels display the corresponding residuals to demonstrate accurate predictions even for early fusion.   }
\label{fig:fig_earlyfusion}
\end{figure*}

\paragraph{Early Fusion Results. }
Early fusion validation results are summarized in Figure \ref{fig:fig_earlyfusion}.
After the transient period (around 10,000 CPT steps), the early fusion runs ($\Dpt=0$) are well predicted by the scaling law.


\paragraph{More Results on the Three-Modality Mixtures.}
We give the full results of the three-modality mixtures in Table \ref{tab:threemodal}.
\begin{table*}[h]
  \centering
  \small
  \setlength{\tabcolsep}{4pt}
  \begin{tabular}{c ccc cc cc}
  \toprule
   \multicolumn{4}{c}{Mixture shares} & \multicolumn{2}{c}{Caption residual} & \multicolumn{2}{c}{Text residual} \\
  \cmidrule(lr){1-4} \cmidrule(lr){5-6} \cmidrule(lr){7-8}
  $f$ & $s_{\mathrm{txt}}$ & $s_{\mathrm{cap}}$ & $s_{\mathrm{int}}$
      & int.\ $=$ text & $w_{\mathrm{int}} = 0.18$
      & int.\ $=$ caption & int.\ $=$ text \\
  \midrule
  \multicolumn{8}{l}{\textit{Caption-free mixtures (text + interleaved)}} \\
  1.00 & 0.85 & 0.00 & 0.15 & -- & -- & $-0.012$ & $+0.007$ \\
  1.00 & 0.60 & 0.00 & 0.40 & -- & -- & $-0.040$ & $+0.024$ \\
  1.00 & 0.40 & 0.00 & 0.60 & -- & -- & $-0.077$ & $+0.048$ \\
  \midrule
  \multicolumn{8}{l}{\textit{Low interleaved fraction ($f \le 0.25$): no new parameter needed for caption}} \\
  0.15 & 0.00 & 0.85 & 0.15 & $-0.004$ & $-0.002$ & -- & -- \\
  0.22 & 0.38 & 0.48 & 0.14 & $-0.016$ & $-0.011$ & $-0.025$ & $+0.023$ \\
  \midrule
  \multicolumn{8}{l}{\textit{High interleaved fraction ($f > 0.25$): one exchange rate recovers most caption runs}} \\
  0.27 & 0.45 & 0.40 & 0.15 & $-0.026$ & $-0.019$ & $-0.022$ & $+0.020$ \\
  0.33 & 0.40 & 0.40 & 0.20 & $-0.029$ & $-0.020$ & $-0.034$ & $+0.027$ \\
  0.40 & 0.00 & 0.60 & 0.40 & $-0.028$ & $-0.019$ & -- & -- \\
  0.50 & 0.20 & 0.40 & 0.40 & $-0.039$ & $-0.022$ & $-0.111$ & $+0.073$ \\
  0.67 & 0.40 & 0.20 & 0.40 & $-0.006$ & $+0.046$ & $-0.057$ & $+0.042$ \\
  0.75 & 0.20 & 0.20 & 0.60 & $-0.057$ & $+0.014$ & $-0.140$ & $+0.082$ \\
  \bottomrule
  \end{tabular}
  \caption{\textbf{Evaluation residuals (observed minus predicted in nats) for three-modality training mixtures (text, caption, interleaved) across varying interleaved multimodal shares ($f$).}  
  Comparisons are made treating interleaved data as text or assigning an effective caption exchange rate ($w_{int} = 0.18$) on the caption loss axis, and treating interleaved data as text or caption on the text loss axis.}
  \label{tab:threemodal}
\end{table*}

\section{Extension to Learning Rate Annealing}
\label{app:anneal}
The law was thus far fitted at a flat learning rate.
Here, we extend it to describe learning rate decay/annealing dynamics.
The result is an additive term to the scaling law using the annealing area $S_2^{\mathrm{cpt}}$ \citep{tissue2024annealing,wang2025learningdynamicscontinualpretraining}:
\begingroup\small
\begin{equation}\label{eq:s2}
\begin{aligned}
L &= L^{\mathrm{flat}} \textcolor{red}{- H N^{-\zeta} g(R)\, S_2^{\mathrm{cpt}}},\\
S_2^{\mathrm{cpt}} &= \sum_i m_i, \quad m_i = \lambda\, m_{i-1} + (\theta_{i-1} - \theta_i), \quad \lambda = 0.996,
\end{aligned}
\end{equation}
\endgroup
with $\theta$ the learning rate in units of its peak.
$g(R) = 1 + b\,\rho(R)$ with $\rho(R) = kR/(kR + 1 - R)$ and $k$ from the flat law (Table~\ref{tab:params}).
On the text axis, $\zeta < 0$, meaning that while text-scarce regimes ($R\to 1$) increase steeply the text loss during flat pre-training, annealing recovers a substantial part of this deficit.
Conversely, on the caption axis, annealing gains are size-invariant ($\zeta \approxeq 0$) but diminish mildly as $R\to 1$ ($b < 0$).
We further find that warmup and annealing effects in PT are negligible.
Note that annealing does not affect the fusion trade-off conclusion, though it makes the loss difference between early/late fusion smaller.

\section{The Chemical Kinetics Analogy}
\label{app:reaction}
We model the shift term in Equation \ref{eq:unified} via the (pseudo) reversible first-order reaction kinetics.
Particularly, we consider heterogeneous surface catalysis, where a continuous external gas stream (training data) reacts with a fixed solid catalyst surface (model parameters).

The model's fixed representational capacity acts as surface sites in either a text-aligned state ($A$) or a multimodal-aligned state ($B$), such that $[A] + [B] = 1$ by conservation laws.
Data are pumped in as a stream containing a molar fraction $R$ of multimodal tokens ($M$) and $1 - R$ of text tokens ($T$).

 Transitions occur when the gas hits the surface:
\begin{align*}
    A + M &\xrightarrow{\kappa_M} B  \\
    B + T &\xrightarrow{\kappa_T} A
\end{align*}
where $\kappa_M$ and $\kappa_T$ are the intrinsic reactivity.
 Since the stream fractions $[M]=R$ and $[T]=1-R$ are externally fixed, mass conservation yields a pseudo-first-order differential equation for state $B$:
\begin{equation}
\begin{aligned}
    \frac{d[B]}{dt} &= \kappa_M R [A] - \kappa_T (1 - R) [B] \\
    &= \kappa_M R - \Big[ \kappa_M R + \kappa_T (1 - R) \Big] [B]
\end{aligned}
\end{equation}

 At dynamic equilibrium ($\frac{d[B]}{dt} = 0$), the steady-state fraction $[B]_\infty$ is:
\begingroup\small
\begin{equation}
    [B]_\infty = \frac{\kappa_M R}{\kappa_M R + \kappa_T (1 - R)} = \frac{k R}{k R + 1 - R} \equiv \rho(R)
\end{equation}
\endgroup
where $k \equiv \kappa_M / \kappa_T$.

Integrating from $[B](0) = 0$ yields $[B](t) = [B]_\infty ( 1 - e^{-\lambda t} )$, where
\begingroup\small
\begin{equation}
    \lambda = \kappa_M R + \kappa_T (1 - R) = \kappa_T \Big( k R + 1 - R \Big)
\end{equation}
\endgroup
Defining the baseline text timescale $\tau \equiv 1/\kappa_T$, $\lambda t$ may be written as $u(t) = ( k R + 1 - R ) \frac{t}{\tau}$.

Substituting $D_{cpt}$ for time $t$, and replacing the exponential decay $1 - e^{-u}$ with a power-law form $1 - (1 + u)^{-\beta}$ to account for the multi-scale spectrum of neural network training (see, e.g., \cite{hutter2021learning}), we directly recover the shift term.
This analogy serves as a toy model to understand and motivate the formulation of the proposed scaling law.

\section{More on Base VLM Evaluation}
\label{app:eval}

\subsection{Open-Ended Document Visual Question Answering}
As mentioned in the main text, evaluating the reference answer's perplexity works partially for DocVQA.
Let us provide a more detailed explanation on this.

For open-ended document visual question answering where candidate reformulation is impractical, we adapt DocVQA to zero-shot, teacher-forced reference perplexity. 
Conditioned on the document image and prompt, we compute the byte-normalized cross-entropy over the ground-truth answer continuation. This formulation provides a direct multimodal analogue to text language modeling perplexity.

\begin{figure*}[t]
\centering
\includegraphics[width=\textwidth]{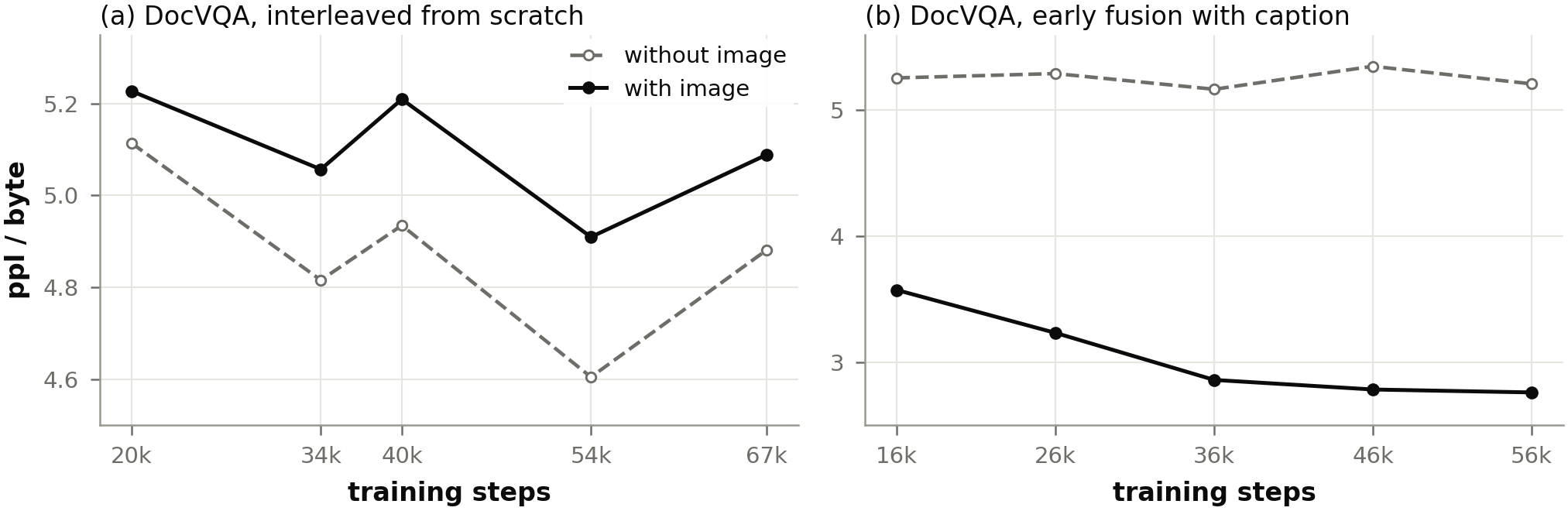}
\caption{\textbf{DocVQA evaluation results via continuation perplexity per byte of the reference answers, with and without the image, for 1B model trained from with interleaved data only (left) and early fusion with caption (right).} We see that only early fusion with caption produces scores that improve with training and positive visual grounding gap. Training with interleaved data even has negative visual grounding gap, raising questions about the effectiveness of interleaved data in multimodal pre-training.}
\label{fig:downstream_trajectory}
\end{figure*}

This works partially: for caption-based pre-training, the score improves with training; when pre-trained with interleaved data from scratch, the visual grounding gap is negative (Figure \ref{fig:downstream_trajectory}).
This suggests that DocVQA is sensitive to pre-training domain bias and serves as an imperfect evaluation proxy for base VLMs; nevertheless, we report it in the paper for completeness as a representative open-ended generation task.
\subsection{Rewriting Strategy}
In our rewriting pipeline, we use declarative templates that are generated deterministically from the question and candidate text to ensure that they share identical syntax to avoid confounders.

For MME, each image comes with a balanced Yes and No pair of questions that differ only in the probed attribute.
We convert both questions to declarative statements using rule-based grammar transformations without inserting explicit negation.
Discarding pairs that fail parsing or lacking matching syntax, we yield 1,158 pairs (2,316 statement) to this end.

For GQA's multi-choice questions, we extract a single statement and slot the each candidate option into it.
We avoid negation in the statement, and hence exclude most binary questions containing negation.
We also drop options that are uninformative like ``\textit{none of the above}''.
After grammatical verification, we are left with 6,405 question sets and 24,985 statements in total.

\subsection{Evaluating Rewritten Benchmarks with Open Models}

\begin{wraptable}{l}{0.55\textwidth}
\centering
\footnotesize
\setlength{\tabcolsep}{3.2pt}
\begin{tabular}{@{}l ccc ccc@{}}
\toprule
& \multicolumn{3}{c}{\textbf{MME (Acc.)}} & \multicolumn{3}{c}{\textbf{GQA (Acc.)}} \\
\cmidrule(lr){2-4} \cmidrule(lr){5-7}
\textbf{Model} & \textbf{Gen.} & \textbf{Ours} & \textbf{Gap ($\Delta$)} & \textbf{Gen.} & \textbf{Ours} & \textbf{Gap ($\Delta$)} \\
\midrule
Qwen2.5-VL-3B & 0.852 & 0.856 & +0.244 & 0.774 & 0.618 & +0.324 \\
Gemma-3-4B-IT & 0.786 & 0.844 & +0.244 & 0.624 & 0.567 & +0.285 \\
SmolVLM2-2.2B & 0.768 & 0.774 & +0.172 & 0.697 & 0.607 & +0.299 \\
\bottomrule
\end{tabular}
\caption{\textbf{Comparison of standard zero-shot evalution versus log-likelihood based evaluation on rewritten benchmarks across open-weight VLMs.} ``Generation'' evaluates letter MCQ selection. ``Ours'' evaluates and ranks byte-normalized log-likelihood of rewritten statements; Gap ($\Delta$) measures the accuracy difference of evaluation with/without image prefix, confirming consistent visual grounding across models.}
\label{tab:open_model_eval}
\end{wraptable}

We further evaluate our rewritten benchmarks with open models, namely Gemma-3-4B-IT \citep{team2025gemma}, Qwen2.5-VL-3B \footnote{\url{https://qwen.ai/blog?id=qwen2.5-vl}}, and SmolVLM2-2.2B \footnote{\url{https://huggingface.co/blog/smolvlm2}}.
Additionally, we evaluate the original MME and GQA benchmarks with the same open models, by prompting them to answer the original questions directly.
Note that these are instruction-tuned models where standard zero-shot generation is feasible.

Results are summarized in \cref{tab:open_model_eval}.
We see that rewriting MME benchmarks can even improve the accuracy of open models, likely due to the more declarative and less ambiguous nature of the rewritten statements from binary Yes/No questions.
Rewriting GQA benchmarks, however, leads to a drop in accuracy, yet still far above chance, reflecting the stricter criteria of full-sentence log-likelihood discrimination compared to predicting the correct option from a set of candidates.
Furthermore, the relative performance ranking of the models remains the same across both formats although the scores differ. 
This indicates that the rewritten benchmarks still evaluate the same underlying visual capabilities.
Importantly, all models maintain large, positive visual grounding gaps.
These results validate the use of our rewritten benchmarks as a proxy for evaluating base VLMs.

\section{More on Fusion Timings}
\label{app:fusion}
\paragraph{Derivation of the Optimal Fusion Timings.}
Under the budget constraint, $\Dcpt$ and $R$ are completely determined by the variable $\Dpt$:
$$D_{\text{cpt}} = (T_{\text{txt}} - D_{\text{pt}}) + T_{\text{cap}}, \qquad R = \frac{T_{\text{cap}}}{D_{\text{cpt}}}$$

The base term can be rewritten as 
\begin{align}
\mathcal{D}_{\mathrm{base, cap}}  &= w_{\rm cap}\,\Dpt + \bigl(R + w_{\rm cap}(1 - R)\bigr)\,\Dcpt \nonumber\\
&= w_{\rm cap}\,T_{\mathrm{txt}} + T_{\mathrm{cap}}
\label{eq:budget}
\end{align}
which holds similarly for the text base term.
Therefore, the base term is constant and drops out of the optimization.

We are left with the shift term.
 Using the identity $(1 - R)D_{\text{cpt}} = T_{\text{txt}} - D_{\text{pt}}$, we express the effective scale factor $X(D_{\text{pt}})$ directly as a linear function of $D_{\text{pt}}$:$$X(D_{\text{pt}}) \equiv (k_{\rm cap} R + 1 - R) D_{\text{cpt}} = k_{\rm cap}\, T_{\text{cap}} + (T_{\text{txt}} - D_{\text{pt}})$$

 Notice that $\rho(R) = \frac{k_{\rm cap} R D_{cpt}}{X(D_{pt})} = \frac{k_{\rm cap} T_{\rm cap}}{X(D_{pt})}$. 
 The caption shift term becomes:
$$\Delta L_{\text{cap}}(D_{\text{pt}}) = -B_{0,\text{cap}}(N) \cdot \frac{k_{\text{cap}} T_{\text{cap}}}{X(D_{\text{pt}})} \left[ 1 - \left( 1 + \frac{X(D_{\text{pt}})}{\tau_{\text{cap}}} \right)^{-\beta_{\text{cap}}} \right]$$
Defining the normalized scale parameter $z(D_{\text{pt}}) = \frac{X(D_{\text{pt}})}{\tau_{\rm cap}} > 0$, we then have
$$\Delta L_{cap}(D_{pt}) = -B_{0,\text{cap}}(N) \frac{k_{\text{cap}} T_{\rm cap}}{\tau_{\rm cap}} \cdot \phi(z(D_{pt})), $$
where $\phi(z) = \frac{1 - (1 + z)^{-\beta}}{z}$.

One can show that $\phi(z)$ is strictly decreasing in $z$ for $\beta > 0$,
because $$\phi'(z) = \frac{\left(1 + (\beta_{\rm cap} + 1)z\right)(1 + z)^{-\beta_{\rm cap} - 1} - 1}{z^2} < 0$$

Therefore the magnitude $|\Delta L_{cap}|$ is strictly increasing with $D_{\text{pt}}$.
The optimal solutions thus lie at the edge:
\begin{align}
\argmin_{\Dpt} L_{\text{cap}} &= T_{\text{txt}} \implies D_{\text{cpt}} = T_{\rm cap}, \; R = 1  \\
\argmin_{\Dpt} L_{\text{txt}} &= 0 \nonumber \\
& \implies D_{\text{cpt}} = T_{\text{txt}} + T_{\text{cap}}, \; R = \frac{T_{\text{cap}}}{T_{\text{txt}} + T_{\text{cap}}}
\end{align}
In other words, early fusion favors lower text loss while late fusion favors lower caption loss.

\paragraph{More on Fusion Timing Experimental Results.}
We give the full results of the fusion timing experiments in \cref{tab:downstream270m,tab:downstream1b}.
We note that S5 reuses our nearest available checkpoint at $\Dpt=$ 32,000 steps to match the target multimodal budget (25,000 caption steps). This leaves the text portion slightly over-budget (+3.3\%, 5,7000 steps versus 56,000 steps for other runs). Given that text tokens contribute minimally to caption loss, and the text performance collapses catastrophically under pure-caption training, this has no measurable impact on the timing's conclusions.
We further note that we do not run S2 for the 1B model to save computational resources, and that the difference with S1 and S3 is of noise-level significance.

\begin{table*}[t]
\centering
\footnotesize
\setlength{\tabcolsep}{4.2pt}
\begin{tabular}{@{}lccclcccccc@{}}
\toprule
& & & & \multicolumn{2}{l}{val loss (nats)} & \multicolumn{5}{l@{}}{text (lm-eval)} \\
\cmidrule(lr){5-6}\cmidrule(lr){7-11}
Schedule & $\Dpt$ & $\Dcpt$ & $R$ & caption & text & HSwag & PIQA & ARC-e & WGrande & LAMB \\
\midrule
S0 & 0 & 56k & 0.447 & 1.1488 & 3.4095 & 0.3691 & 0.6529 & 0.4933 & 0.5217 & 69.8 \\
S1 & 4k & 52k & 0.482 & 1.1418 & 3.4194 & 0.3668 & 0.6436 & 0.4912 & 0.5146 & 73.0 \\
S2 & 8k & 48k & 0.522 & 1.1277 & 3.4286 & 0.3678 & 0.6235 & 0.4928 & 0.5036 & 73.7 \\
S3 & 16k & 40k & 0.626 & 1.0983 & 3.4574 & 0.3613 & 0.6349 & 0.4836 & 0.5367 & 89.3 \\
S4 & 24k & 32k & 0.782 & 1.0486 & 3.5341 & 0.3591 & 0.6360 & 0.4726 & 0.5162 & 80.0 \\
S5  & 32k & 25k & 1.000 & 0.9902 & 6.3135 & 0.2938 & 0.5871 & 0.3215 & 0.5130 & 2466.4 \\
\midrule
\multicolumn{6}{@{}l}{Spearman $r_s$ } & $-0.94$ & $-0.77$ & $-0.94$ & $-0.20$ & $+0.94$ \\
\bottomrule
\end{tabular}

\medskip

\begin{tabular}{@{}lcccccc@{}}
\toprule
& \multicolumn{4}{l}{likelihood ranking (accuracy, with image)} & \multicolumn{2}{l@{}}{generation ppl} \\
\cmidrule(lr){2-5}\cmidrule(lr){6-7}
Schedule & MME & gap & GQA & gap & DocVQA & $\Delta$ \\
\midrule
S0 & 0.6701 & +0.0941 & 0.4103 & +0.1280 & 3.88 & +1.98 \\
S1 & 0.6796 & +0.1123 & 0.3694 & +0.0924 & 3.73 & +1.96 \\
S2 & 0.6788 & +0.1097 & 0.3975 & +0.1198 & 3.79 & +2.10 \\
S3 & 0.6779 & +0.0959 & 0.3891 & +0.1085 & 4.13 & +1.75 \\
S4 & 0.6883 & +0.1157 & 0.3969 & +0.1226 & 3.78 & +2.20 \\
S5 & 0.6770 & +0.0976 & 0.3869 & +0.1027 & 3.95 & +6.11 \\
\midrule
\multicolumn{1}{@{}l}{$r_s$ } & $-0.20$ & $-0.31$ & $+0.37$ & $+0.26$ & $-0.31$ & $-0.60$ \\
\bottomrule
\end{tabular}
\caption{\textbf{Downstream benchmark performance and final validation losses for the fusion timing experiments on the 300M model.} 
Vision benchmarks are evaluated using our rewriting strategy (MME and GQA) or via continuation perplexity (DocVQA), along with the gap between evaluation with and without image prefix. The bottom rows report Spearman rank correlation $r_s$ between downstream metrics and their corresponding validation loss across schedules S0-S5.}
\label{tab:downstream270m}
\end{table*}

\begin{table*}[t]
\centering
\footnotesize
\setlength{\tabcolsep}{4.2pt}
\begin{tabular}{@{}lccclcccccc@{}}
\toprule
& & & & \multicolumn{2}{l}{val loss (nats)} & \multicolumn{5}{l@{}}{text (lm-eval)} \\
\cmidrule(lr){5-6}\cmidrule(lr){7-11}
Schedule & $\Dpt$ & $\Dcpt$ & $R$ & caption & text & HSwag & PIQA & ARC-e & WGrande & LAMB \\
\midrule
S0 & 0 & 56k & 0.447 & 0.9659 & 3.0258 & 0.4862 & 0.7029 & 0.5711 & 0.5264 & 27.0 \\
S1 & 4k & 52k & 0.482 & 0.9600 & 3.0360 & 0.4795 & 0.6921 & 0.5720 & 0.5517 & 27.3 \\
S3 & 16k & 40k & 0.626 & 0.9304 & 3.0791 & 0.4702 & 0.6997 & 0.5787 & 0.5438 & 24.8 \\
S4 & 24k & 32k & 0.782 & 0.9029 & 3.1589 & 0.4569 & 0.6915 & 0.5673 & 0.5391 & 32.3 \\
S5 & 32k & 25k & 1.000 & 0.8705 & 5.8537 & 0.3144 & 0.6262 & 0.3611 & 0.5130 & 557.6 \\
\midrule
\multicolumn{6}{@{}l}{Spearman $r_s$ } & $-1.00$ & $-0.90$ & $-0.60$ & $-0.40$ & $+0.70$ \\
\bottomrule
\end{tabular}

\medskip

\begin{tabular}{@{}lcccccc@{}}
\toprule
& \multicolumn{4}{l}{likelihood ranking (accuracy, with image)} & \multicolumn{2}{l@{}}{DocVQA ppl} \\
\cmidrule(lr){2-5}\cmidrule(lr){6-7}
Schedule & MME & gap & GQA & gap & DocVQA & $\Delta$ \\
\midrule
S0  & 0.7064 & +0.1131 & 0.4134 & +0.1386 & 3.10 & +2.35 \\
S1 & 0.7055 & +0.1097 & 0.4236 & +0.1429 & 2.91 & +2.30 \\
S3 & 0.7116 & +0.1218 & 0.4158 & +0.1374 & 2.94 & +2.37 \\
S4 & 0.7090 & +0.1278 & 0.4159 & +0.1326 & 2.67 & +2.64 \\
S5 & 0.6839 & +0.1088 & 0.4209 & +0.1318 & 2.80 & +5.03 \\
\midrule
\multicolumn{1}{@{}l}{$r_s$ } & $+0.20$ & $+0.10$ & $-0.40$ & $+0.90$ & $+0.80$ & $-0.90$ \\
\bottomrule
\end{tabular}
\caption{\textbf{Downstream benchmark performance and final validation losses as in Table \ref{tab:downstream270m} but evaluated for the 1B model.}}
\label{tab:downstream1b}
\end{table*}

\end{document}